\documentclass[12pt,a4paper,harvard]{article}

\usepackage[utf8]{inputenc}
\usepackage{mathptmx}
\usepackage{setspace}
\usepackage{makecell}
\usepackage{multibib}
\usepackage{graphicx,xcolor,comment}
\usepackage{pgfplots}
\pgfplotsset{compat=1.7}
\usepackage{dingbat}
\usepackage{multirow}
\usepackage{natbib}
\usepackage{lscape}
\usepackage{authblk}
\usepackage{hyperref}
\usepackage{cleveref}
\usepackage{multicol}
\usepackage{adjustbox}
\usepackage{geometry}
\usepackage{caption}
\usepackage{rotating} 
\usepackage{float}

\usepackage{subcaption}
\usepackage{booktabs}
\newgeometry{left=0.8in,right = 0.8in} 
\usepackage{csquotes}
\providecommand{\keywords}[1]
{
  \small	
  \textbf{\textit{Keywords: }} #1
}
\title{A Review of AI and Machine Learning Contribution in Predictive Business Process Management (Process Enhancement and Process Improvement Approaches)}

\author[a]{Mostafa Abbasi }
\author[c]{ Rahnuma Islam Nishat\footnote{The work was done during the author's postdoctoral fellowship at UBC Okanagan.} }
\author[b]{ Corey Bond }
\author[b]{ John Brandon Graham-Knight }
\author[b]{Patricia Lasserre}
\author[b]{ Yves Lucet }
\author[a]{ Homayoun Najjaran\footnote{Corresponding author, najjaran@uvic.ca} }
\affil[a]{Faculty of Engineering and Computer Science, University of Victoria, BC, Canada}
\affil[b]{Computer Science, Irving K. Barber Faculty of Science, University of British Columbia, Kelowna, BC, Canada}
\affil[c]{Computer Science, Brock University, St. Catharines, Canada}

\usepackage{listingsutf8}

\begin{document}

\maketitle

\begin{abstract}
\textbf{Purpose-} The significance of business processes has fostered a close collaboration between academia and industry. Moreover, the business landscape has witnessed continuous transformation, closely intertwined with technological advancements. Our main goal is to offer researchers and process analysts insights into the latest developments concerning Artificial Intelligence (AI) and Machine Learning (ML) to optimize their processes in an organization and identify research gaps and future directions in the field.

\textbf{Design/methodology/approach-} In this study, we perform a systematic review of academic literature to investigate the integration of AI/ML in business process management (BPM). We categorize the literature according to the BPM life-cycle and employ bibliometric and objective-oriented methodology, to analyze related papers.

\textbf{Findings-} In business process management and process map, AI/ML has made significant improvements using operational data on process metrics. These developments involve two distinct stages: (1) process enhancement, which emphasizes analyzing process information and adding descriptions to process models, and (2) process improvement, which focuses on redesigning processes based on insights derived from analysis.

\textbf{Research limitations/implications-} While this review paper serves to provide an overview of different approaches for addressing process-related challenges, it does not delve deeply into the intricacies of fine-grained technical details of each method. This work focuses on recent papers conducted between 2010 and 2024. 

\textbf{Originality/value-} This paper adopts a pioneering approach by conducting an extensive examination of the integration of AI/ML techniques across the entire process management life-cycle. Additionally, it presents groundbreaking research and introduces AI/ML-enabled integrated tools, further enhancing the insights for future research.

\end{abstract}

\section{Introduction and Theoretical Background}

The efficient operation of a corporation relies on the seamless integration of people and business processes to obtain desired outcomes. Business Process Management (BPM) is the main element to regulate and govern this integration because boosting efficiency in any organization often necessitates changing its processes in accordance with an informed perspective~\citep{powell2001measurement}. BPM can be described in numerous ways but to be concise and clear, it involves the analysis and diagnosis of process shortcomings, the redesign of the process, and the implementation of corrective measures~\citep{Luo2011}.  Analysis and diagnosis share a common essence, as both require a comprehensive study and data analysis to identify deficiencies and develop a roadmap. Redesign and implementation involve tangible actions that actively focus on how to rearticulate and rearrange a process in terms of its dependent and independent tasks and resources based on the analysis~\citep{Mustansir2022}. Once the process evaluation and planning are established, process improvement can be pursued~\citep{Urh2019}. 

In summary, BPM consists of two main components: (1) process enhancement, which involves monitoring and evaluating the current process to diagnose inefficiencies, and (2) process improvement, which includes redesigning the process and implementing corrective actions based on the preceding analysis. For clarity, in the scholarly literature, the term ``predictive business process monitoring'' is commonly used to describe process enhancement through the application of predictive analytics to business processes based on historical data. Process enhancement focuses on predicting various process perspectives, such as time, data, resources, costs, and quality, for ongoing process instances~\citep{DiFrancescomarino2018}. 
However, despite its importance, a coherent and standardized consensus regarding the precise application of these terms in the scholarly literature remains elusive due to the variation in terminology~\citep{vanderAalst2022}. While process monitoring is valuable for tracking and gathering information, it may not provide actionable recommendations for process improvement. Therefore, to achieve a more comprehensive analysis, integrating process evaluation is essential~\citep{deLeoni2022,DiFrancescomarino2022}. The deliberate usage of ``process enhancement'' underscores a purposeful emphasis on the analysis of processes and their outcomes to both extend and improve them. In the following sections, we will introduce a set of reviews on process improvement and process enhancement to describe the research area within these fundamental aspects of Business Process Management (BPM).

\subsection{Process Improvement}

Many surveys seek to find solutions to handle challenges including insufficient BPM knowledge in education~\citep{Seethamraju2012}, in IT sector~\citep{Alotaibi2017} as a critical enabler in process improvement. Change management, resulting from the implementation of process improvement methods, presents an additional challenge in this context.~\citet{Sujova2018} investigated the impact of various methods and also provided an overview of how businesses dealt with the transitions in the process.

According to the reviews, there is a variety of factors including people and tools that contribute to the success of a BPI  project. Process owners as a prominent part of process improvement methods have been investigated in~\citep{Danilova2019} , specifically with regard to elucidating stakeholders' roles, responsibilities, and the required characteristics of the team to have a successful improvement strategy.  Also,~\citet{Arias2018} placed emphasis on human resources within the context of BPM and thus they evaluated papers to identify  the approaches, business drivers, issues, and challenges in human resource allocation. In addition, ~\citet{Pourmirza2017} believes BPMS (business process management system) is a crucial element in achieving excellence in business process management. Therefore, they reviewed around 600 papers to classify and analyze the different BPMS architectures and eventually, they introduced 41 different architectures that could be applicable in this context. Furthermore, the significance of culture has been underscored by~\citet{VomBrocke2011} as a facilitator for ensuring the effectiveness of employing the BPI method in an organization. They have elaborated the frame of reference regarding culture’s role and influences through their scholarly works. Besides, surveys dedicated attention to the exploration of digital innovation~\citep{Ahmad2020} and blockchain~\citep{Viriyasitavat2022, info15010009} as an advantageous tool.

During the past few decades, there has been a noticeable change in methodologies, coinciding with the big data age and developments in databases. These advances have made it much easier for improvement teams to provide appropriate and meaningful corrective actions\citep{vera2013business}. As a result, process mining has developed into an active area that stands between data mining, machine learning, and the modeling and analysis of business processes. Its goal is to identify, track, and enhance actual business processes by extracting knowledge from event logs. In this area, thorough examinations were conducted by~\citet{van2012process} and later expanded upon by~\citep{SantosGarcia2019Processminingtechniques} The authors have thoroughly elaborated on existing techniques and approaches within process mining, and  systematically categorized the literature into distinct groups and delivered an extensive overview of the field.

\subsection{Process Enhancement}

Process enhancement is one of the three main attributes of process mining (\emph{i.e.}, (a) discovery, (b) conformance checking, and (c) enhancement).  If the process model is provided this capability can be used to modify or extend the existing model. Extending a process model with performance data, identifying bottlenecks, or detecting deviations are examples of this capability. The insights obtained in this step fuel the suggestions of business process analysts and process owners to change the process. Finally, process enhancement can be used to track processes over time and assist in continuous improvement~\citep{Aalst2011}. 
On the other hand, process enhancement specifically provides process improvement recommendations based on the metrics (\emph{e.g.}, remaining process time, waiting time) and helps with the root-cause analysis and according to this information using event logs, improvements or changes can be made. On the other hand, BPI (business process improvement) consists of techniques and approaches (\emph{e.g.}, Six Sigma, Lean, etc.)  to increase the effectiveness and efficiency of the actual business processes that provide output to internal and external customers. For this, there are various techniques to find improvement opportunities and implement corrective actions. Process enhancement is one of the methods that play a role in introducing improvement opportunities and developing the roadmap for BPI~\citep{Aalst2016DataScienceAction}.

It should be noted that this paper focuses on organizations that have modeled their own processes (current situation); however, they aim to improve their business processes derived from the event logs and process data. To be more specific, different process properties, including the goals of the process, the desired output, and the current process model, have been developed. However, the business wants to find and implement corrective actions to enhance efficiency. 

Process enhancement is a relatively nascent area, resulting in a more limited collection of review papers compared to the longer-standing area of process improvement.  However, scholars have demonstrated the capacity of process enhancement across various applications and industries. In particular,~\citep{rojas2016process,williams2018process} conducted a comprehensive review of scholarly papers on healthcare processes, highlighting specific approaches within this crucial sector.

Goal-oriented process enhancement is another area of interest among researchers, particularly in reviews focusing on goals such as addressing concept drift~\citep{sato2021survey} to maintain process orientation, optimize duration, and improve resource utilization~\citep{guzzo2022process}. Moreover, \citet{lopes2023assessing} proposed a comprehensive framework called bPERFECT (Business Process Evaluation and Research Framework for Enhancement and Continuous Testing) to discuss testing and verification approaches in business process models. These approaches, which address the problem from different perspectives and with varying degrees of automation, process complexity, and process maturity, have been shown to improve the efficiency of process enhancement methods.

As discussed earlier, identifying solutions for challenges in business process management is crucial for organizations to effectively apply methods and techniques in a cost-efficient manner. One significant obstacle in process enhancement is data cleaning, which can hinder progress. To address this issue,~\citet{marin2021event}  reviewed preprocessing techniques specifically designed for event logs.

\begin{figure}[ht]
    \centering
         \includegraphics[width=0.85\textwidth]{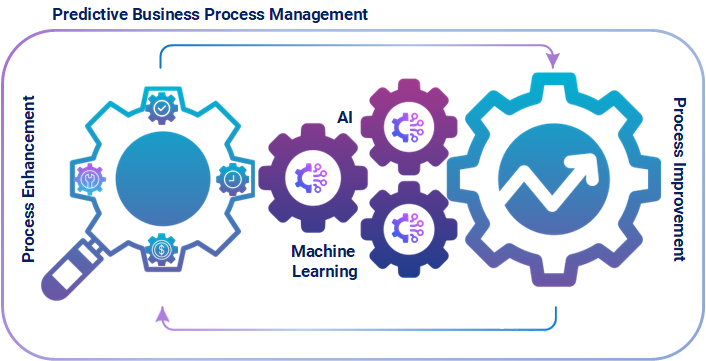}
    \caption{Research Scope$^{1}$}
    \label{fig:Research Scope (ML/AI and PBPM)}
    
\end{figure}

A growing body of research (as summarized in~\Cref{Related Works}) highlights the transformative potential of ML and AI across various industries,
including business process management (BPM). However, a critical gap exists in analyzing the specific impact of these techniques on process improvement and enhancement methodologies. While existing literature extensively explores both process improvement and enhancement, it fails to emphasize the combination of both methodologies. On the other hand, this research addresses the identified gap by focusing on Predictive Business Process Management (PBPM) (~\Cref{fig:Research Scope (ML/AI and PBPM)}). PBPM leverages cutting-edge ML and AI techniques to provide a holistic approach for optimizing business processes. It empowers organizations to not only improve and streamline existing workflows but also proactively evaluate and redesign processes based on valuable insights gleaned from these advanced methods.

\footnotetext[1]{This figure was prepared using images from www.flaticon.com.}

\begin{table}[!ht]
\centering
\captionsetup{skip=3pt}
\caption{Related Works}
\label{Related Works}
\resizebox{500pt}{!}{
\begin{tabular}{cccccccccccccccccc}
\hline
\multirow{2}{*}{Name}       & \multirow{2}{*}{\begin{tabular}[c]{@{}c@{}}Process\\ Improvement\end{tabular}} & \multirow{2}{*}{\begin{tabular}[c]{@{}c@{}}Process\\ Enhancement\end{tabular}} & \multicolumn{2}{c}{\begin{tabular}[c]{@{}c@{}}Methods \& Approach \\ Introduction\end{tabular}} &  & \multicolumn{5}{c}{Actors \&  Tools}                                                                                                                                                                                                                                        &  & \multicolumn{5}{c}{Challenges and Issues}                                                                                                                                                                                                                                  & \multirow{2}{*}{Case Study} \\ \cline{4-5} \cline{7-11} \cline{13-17}
                            &                                                                                &                                                                                & Contributions                                    & Objectives                                   &  & \begin{tabular}[c]{@{}c@{}}Process \\ Owners\end{tabular} & BPMS       & \begin{tabular}[c]{@{}c@{}}Digital \\ Transforamtion\end{tabular} & \begin{tabular}[c]{@{}c@{}}Culture and \\ Team\end{tabular} & \begin{tabular}[c]{@{}c@{}}AI \& \\ Machine Learning\end{tabular} &  & Agility    & \begin{tabular}[c]{@{}c@{}}Insufficient \\ Knowdelge\end{tabular} & \begin{tabular}[c]{@{}c@{}}Change \\ Management\end{tabular} & \begin{tabular}[c]{@{}c@{}}Security \\ Requirements\end{tabular} & \begin{tabular}[c]{@{}c@{}}Data\\ Cleaning\end{tabular} &                             \\ \cline{1-5} \cline{7-11} \cline{13-18} 
\citep{Zellner2011}         & \checkmark                                                                     &                                                                                & \checkmark                                       &                                              &  &                                                           &            &                                                                   &                                                             &                                                                  &  &            & \checkmark                                                        &                                                              & \multicolumn{1}{l}{}                                             & \multicolumn{1}{l}{}                                    & General                     \\
\citep{VomBrocke2011}       & \checkmark                                                                     &                                                                                &                                                  &                                              &  &                                                           &            &                                                                   & \checkmark                                                  &                                                                  &  &            &                                                                   &                                                              & \multicolumn{1}{l}{}                                             & \multicolumn{1}{l}{}                                    & General                     \\
\citep{Seethamraju2012}     & \checkmark                                                                     &                                                                                &                                                  &                                              &  &                                                           &            &                                                                   &                                                             &                                                                  &  &            &                                                                   &                                                              &                                                                  &                                                         & General                     \\
\citep{Segatto2013}         & \checkmark                                                                     &                                                                                &                                                  &                                              &  &                                                           &            &                                                                   &                                                             &                                                                  &  &            &                                                                   &                                                              &                                                                  &                                                         & General                     \\
\citep{UrionaMaldonado2020} & \checkmark                                                                     &                                                                                &                                                  &                                              &  &                                                           &            &                                                                   &                                                             &                                                                  &  &            &                                                                   &                                                              &                                                                  &                                                         & General                     \\
\citep{Mendes2018}          & \checkmark                                                                     &                                                                                &                                                  &                                              &  &                                                           &            &                                                                   &                                                             &                                                                  &  &            &                                                                   &                                                              &                                                                  &                                                         & General                     \\
\citep{rojas2016process}    &                                                                                & \checkmark                                                                     & \checkmark                                       & \checkmark                                   &  &                                                           &            &                                                                   &                                                             &                                                                  &  &            &                                                                   &                                                              &                                                                  &                                                         & Healthcare                  \\
\citep{Alotaibi2017}        & \checkmark                                                                     &                                                                                &                                                  &                                              &  &                                                           &            &                                                                   &                                                             &                                                                  &  & \checkmark & \checkmark                                                        &                                                              & \checkmark                                                       &                                                         & General                     \\
\citep{Pourmirza2017}       & \checkmark                                                                     &                                                                                &                                                  &                                              &  &                                                           & \checkmark &                                                                   &                                                             &                                                                  &  &            &                                                                   &                                                              &                                                                  &                                                         & General                     \\
\citep{Bazan2020}           & \checkmark                                                                     &                                                                                &                                                  &                                              &  &                                                           &            &                                                                   &                                                             &                                                                  &  &            &                                                                   &                                                              &                                                                  &                                                         & General                     \\
\citep{Arias2018}           & \checkmark                                                                     &                                                                                &                                                  &                                              &  &                                                           &            &                                                                   & \checkmark                                                  &                                                                  &  &            &                                                                   &                                                              &                                                                  &                                                         & General                     \\
\citep{Sujova2018}          & \checkmark                                                                     &                                                                                &                                                  &                                              &  &                                                           &            &                                                                   &                                                             &                                                                  &  &            & \checkmark                                                        & \checkmark                                                   &                                                                  &                                                         & General                     \\
\citep{Danilova2019}        & \checkmark                                                                     &                                                                                &                                                  &                                              &  & \checkmark                                                &            &                                                                   & \checkmark                                                  &                                                                  &  &            &                                                                   &                                                              &                                                                  &                                                         & General                     \\
\citep{Ahmad2020}           & \checkmark                                                                     &                                                                                &                                                  & \checkmark                                   &  &                                                           &            & \checkmark                                                        &                                                             &                                                                  &  &            &                                                                   &                                                              &                                                                  &                                                         & General                     \\
\citep{sato2021survey}      &                                                                                & \checkmark                                                                     &                                                  &                                              &  &                                                           &            &                                                                   &                                                             &                                                                  &  &            &                                                                   &                                                              &                                                                  &                                                         & General                     \\
\citep{guzzo2022process}    &                                                                                & \checkmark                                                                     &                                                  &                                              &  &                                                           &            &                                                                   &                                                             &                                                                  &  &            &                                                                   &                                                              &                                                                  &                                                         & Healthcare                  \\
\citep{marin2021event}      &                                                                                & \checkmark                                                                     & \checkmark                                       &                                              &  &                                                           &            &                                                                   &                                                             &                                                                  &  &            &                                                                   &                                                              &                                                                  & \checkmark                                              & General                     \\
\citep{Malinova2022}        & \checkmark                                                                     &                                                                                & \checkmark                                       & \checkmark                                   &  &                                                           &            &                                                                   &                                                             &                                                                  &  &            &                                                                   &                                                              &                                                                  &                                                         & General                     \\
\citep{Viriyasitavat2022}   & \checkmark                                                                     &                                                                                &                                                  &                                              &  &                                                           &            & \checkmark                                                        &                                                             &                                                                  &  &            &                                                                   &                                                              &                                                                  &                                                         & General                     \\
 \citep{lopes2023assessing}& & \checkmark& \checkmark& \checkmark& & & \checkmark& \checkmark& \checkmark& & & \checkmark& & \checkmark& & &General                     \\
 \citep{Tsakalidis_Vergidis_2024}& \checkmark& & & \checkmark& & & \checkmark& & & & & & & & & &General                     \\
This work& \checkmark                                                                     & \checkmark                                                                     & \checkmark                                       & \checkmark                                   &  &                                                           &            &                                                                   & \checkmark                                                  & \checkmark                                                       &  &            &                                                                   & \checkmark                                                   &                                                                  & \checkmark                                              & General                     \\ \hline
\end{tabular}
}
\end{table}

\subsection{Research Questions}

The main purpose of this study was to develop a comprehensive analysis of the academic literature on business process improvement, process enhancement, and integrated methodologies with a specific focus on the application of AI and machine learning techniques.  This paper defines its scope to assist business process experts in identifying organization-specific and feasible Business Process Management (BPM) solutions, particularly in light of recent AI advancements. The aim is to introduce this evolving field to BPM analysts, aiding them in adapting to these emerging technological changes. It is important to introduce this area to BPM analyst to help them to adapt with these changes. Additionally, as businesses increasingly seek integrated solutions for process enhancement and improvement, a growing body of research has emerged to address this need. These papers not only identify process shortcomings but also aim to enhance processes to maximize effectiveness and efficiency. In this review paper, we aim to examine these methods, approaches, and contributions in comparison to traditional models.

Tackling the two aforementioned gaps, we focus on the following research questions.

RQ1: In light of advancements in AI and machine learning in recent years, what is the impact of these developments on process improvement and process enhancement and how do researchers employ AI and machine learning to advance the field of BPM?

RQ2: What approaches have been introduced in the literature that address the integration of process enhancement and process improvement?

RQ3: Are there differences in the manner in which an organization employs process enhancement and process improvement techniques as separate entities compared to when they adopt integrated approaches?

The article is organized as follows. In Section 2, the research methodology is presented and the bibliometric analysis is presented to evaluate and identify the research impacts and trends in this context. Section 3 presents the main process enhancement approaches according to the objectives. The most relevant papers on AI and machine learning methods in process improvement algorithms are presented in Section 4. Section 5 outlines the papers to introduce the combined process enhancement and process improvement approaches to cover one of the research questions. Finally, a conclusion is given in Section 6.

\section{Research design and methodology}

\begin{figure}[H]
    \centering
         \includegraphics[width=1\textwidth]{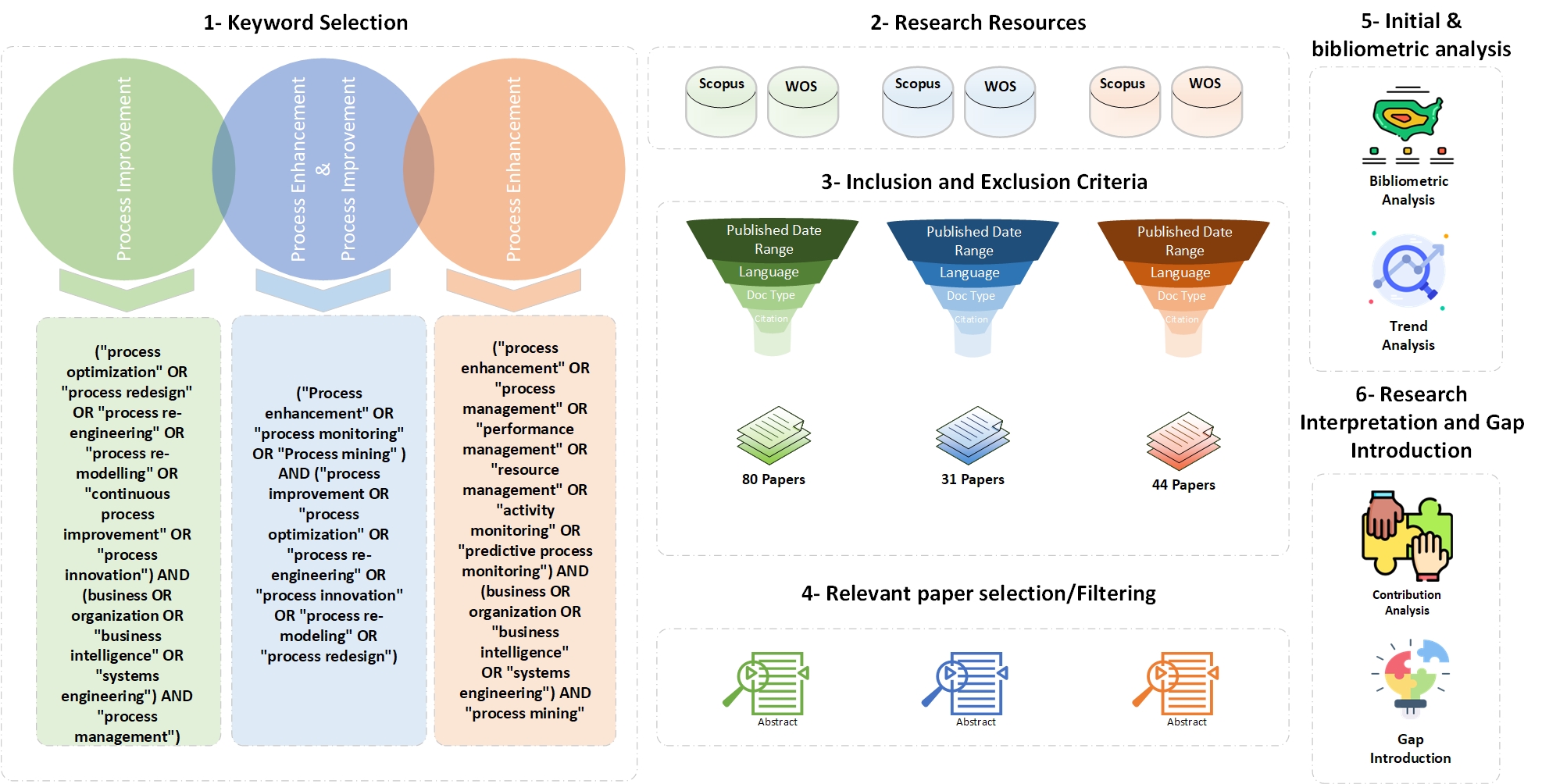}
    \caption{Research Methodology}
    \label{fig:Research Methodology}
\end{figure}

As it is shown in~\Cref{fig:Research Methodology}, this paper introduces a 6-step framework designed to effectively address research questions, achieve desired objectives, and follow systematic research review methodologies, such as PRISMA. The framework was meticulously designed to ensure careful selection of the desired range necessary to answer the question while maintaining precision and accuracy in selecting appropriate articles. The research is comprehensive in nature and involves an extensive literature review in the field, aimed at providing a scientific and reasoned analysis. To avoid ambiguity, the methodology is thoroughly and clearly explained throughout the entire article. The goal of this research is to produce useful and applicable findings and results in both the business world and academic areas.

Furthermore, the overarching goal of the framework is to strategically enhance the efficiency, efficacy, and broader impact of the study in the relevant field. Here is a detailed explanation of the 6-step methodology that follows.

\textbf{1- Keyword Selection:} The initial step of this methodology is to select relevant and suitable words or phrases that precisely convey the key themes and ideas addressed in the research question. In the methodology, we have partitioned the research subject matter into three significant categories including (1) Process Enhancement (2) Process Improvement (3) Process Enhancement and Process Improvement. This systematic approach aids us in identifying, with precision, the articles that correspond to each category from various research resources. It is particularly valuable in categorizing the analysis presented according to the appropriate category. In each category, we have determined relevant keywords, which we will expound upon in \Cref{tab:Keywords}.

\textbf{2- Research Resources:} For this study, we made use of resources including Semantic Scholars, Web of Science, and Scopus. With access to a multitude of scholarly literature from a variety of subjects, all three search engines are potent tools for academic research.

\textbf{3- Inclusion and Exclusion Criteria:} This step confirms that the reviewed studies adhere to the predetermined standards for relevance and quality. The review can offer a more thorough and accurate assessment of the research available on this issue by choosing papers that satisfy these requirements. We have established a set of criteria for this purpose as follows: Document Type:  Journal or Conference Paper, Published Date: 2010 - 2024. Language: English (196 Papers)

\textbf{4- Relevant Paper Selection/Filtering:}  We explored resources suited for academic research to find publications that matched the specified search terms and criteria. To quickly identify relevant research and narrow down our selection in terms of approaches and scope, we initially focused on the abstract and introduction of each paper. It enables us to ascertain the relevant papers that apply machine learning and AI methods. (46 Papers)

\textbf{5- Initial and Bibliometric Analysis:}  A thorough analysis of publication trends and a general investigation within scientific journals was conducted in this study  to determine the contribution to the research area. Moreover, insights into the evolution and impact of business process literature was included in \Cref{Research Results and Bibliometric Analysis}.

\textbf{6- Research  Interpretation and Gap Introduction:}  This step entails evaluating the existing literature critically and examining the advantages, disadvantages, and restrictions of the approaches to enhance and improve business processes in the organization. Also, we discuss gaps in the literature that help to establish the necessity of further study as well as offer a clear direction for future research.

\subsection{Research Results and Bibliometric Analysis}
\label{Research Results and Bibliometric Analysis}
\begin{figure}[htbp]
    \centering
    \includegraphics[width=0.75\textwidth]{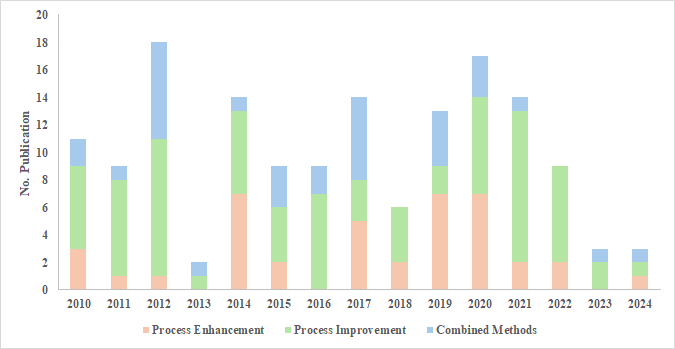}
    \caption{Number of published papers in 2010-2024}
    \label{fig:NO. published paper in 2010-2022}
\end{figure}

The number of publications in each area during the last twelve years (2010-2024) is shown in  \Cref{fig:NO. published paper in 2010-2022},  and contributing journals are introduced in  \Cref{tab:TopTwelveJournals}. The majority of research papers each year is primarily focused on process improvement. This highlights the recent impact of machine learning and AI in innovative solutions for process improvement problems. Nevertheless, there remains a necessity to expand the application of machine learning techniques for process enhancement and to explore integrated approaches.

Nine journals were popular for their contributions over the years. Business Process Management Journal is the highest, with 27 publications during the period 2010–2024 (about 25\% of all articles), followed by Decision Support Systems with 6 publications.

\begin{table}[htbp]
\centering
\begin{minipage}{0.5\textwidth}
  \centering
  \captionsetup{skip=3pt}
  \caption{Top 9 journals contributing to this area}
  \resizebox{250pt}{!}{
  \begin{tabular}{lcr}
    \hline
    Journal Name & No. Publication \\
    \hline
    Business Process Management Journal & 27 \\
    Decision Support Systems & 6 \\
    Information Systems & 3 \\
    Software and Systems Modeling & 2 \\
    Business and Information Systems Engineering & 2 \\
    Enterprise Information Systems & 2 \\
    International Journal of Business Process \\ Integration and Management & 2 \\
    International Journal of System Assurance \\ Engineering and Management & 2 \\
    Journal of Healthcare Management & 2 \\
    Other Journals & 1 \\
    \hline
  \end{tabular}}
  \label{tab:TopTwelveJournals}
\end{minipage}%
\hfill
\begin{minipage}{0.47\textwidth}
  \centering
  \captionsetup{skip=3pt}
  \caption{Keywords}
  \resizebox{250px}{!}{
  \begin{tabular}{lll}
    \hline
    Category & & Keyword \\
    \hline
    Process Enhancement & & \begin{tabular}[c]{@{}l@{}}process management, performance \\ management, resource management,\\ activity monitoring, \\ predictive process monitoring,\\ process mining\end{tabular} \\
    & & \\
    Process
    
    Improvement & & \begin{tabular}[c]{@{}l@{}}process redesign, process optimization, \\ process re-engineering, process re-modeling, \\ continuous process improvement,\\ process innovation\end{tabular} \\
    & & \\
    Business Context & & \begin{tabular}[c]{@{}l@{}}business, organization, business intelligence, \\ systems engineering\end{tabular} \\
    \hline
  \end{tabular}}
  \label{tab:Keywords}
\end{minipage}
\end{table}

Bibliometric analysis serves as a valuable and effective tool for examining research papers, which encompasses various aspects such as authors, their network,  keywords, and the evolution of keywords (trends). In the following, we present the figures accompanied by explanations, aimed at providing practical insights for researchers.Figures were generated using VOS Viewer. 

\begin{figure}[ht]
    \centering
         \includegraphics[width=1.0\textwidth]{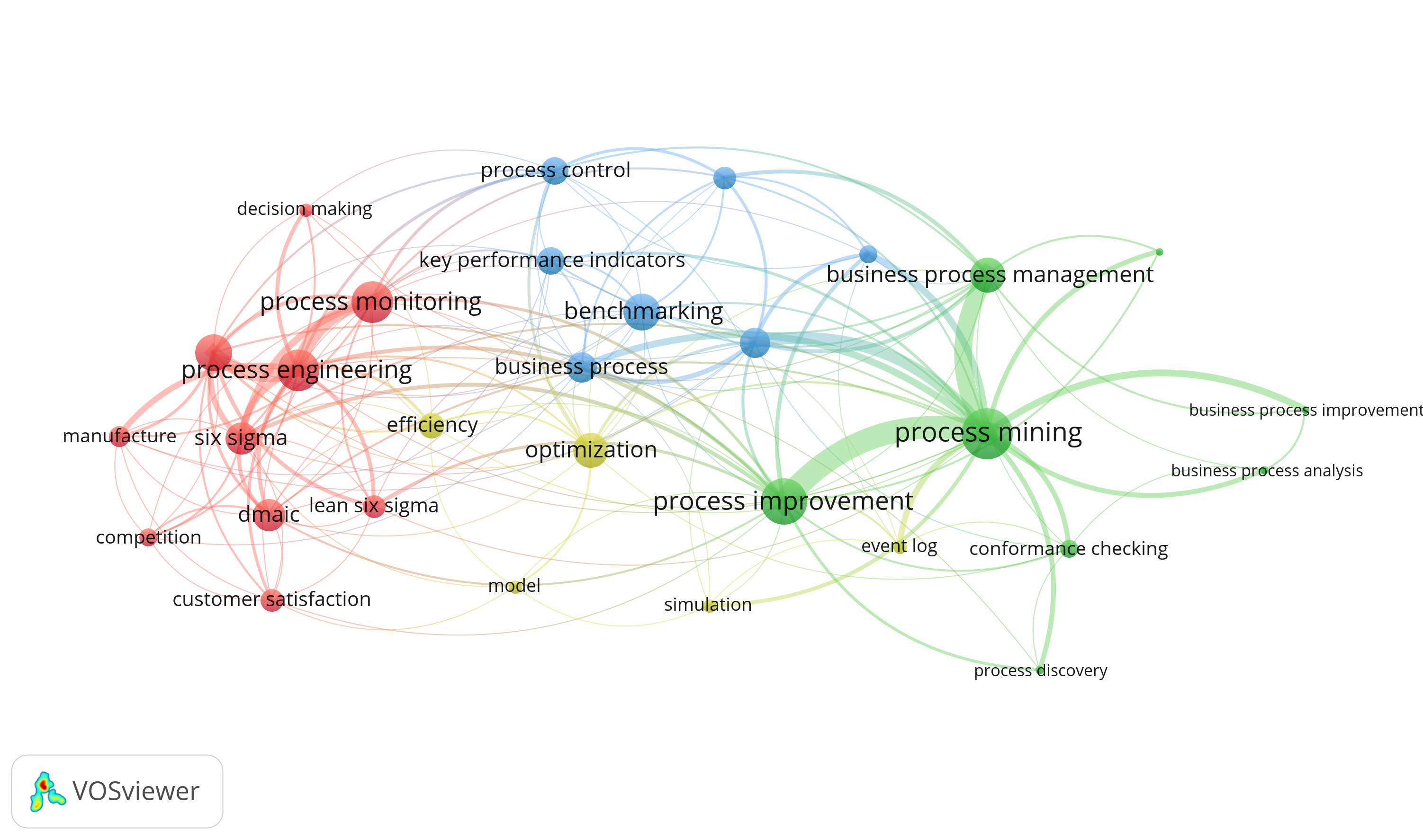}
    \caption{Keyword network links and clusters in this context}
    \label{fig:Keyword-network-Links}
\end{figure}

Keyword analysis yielded 28 keywords that met the specified criteria, namely the minimum number of occurrences of author keywords was four keywords as indicated by the network visualization (\Cref{fig:Keyword-network-Links}). Among these, \textbf{Process Mining, Process Improvement, Process Monitoring, Process Engineering, and Business Process Management} are the five keywords that dominate. Furthermore, the predictive business process management (PBPM) domain encompasses  four key clusters connected to  subjects, each distinguished by a distinct color code.

\Cref{fig:Keyword-network-Links} clearly demonstrates the relation between business process management, process improvement, and process mining techniques as documented in the existing literature. This figure visually represents the proximity and connections among papers categorized under these keywords across various clusters. Cluster 1 is visualized in green and it forms the focal point of this paper. The Second cluster (red) primarily revolves around process improvement approaches and techniques. On the other hand, the blue cluster, representing generic terms in business execution, is interconnected with both the green and red clusters. Notably, the literature on process enhancement, in comparison to process improvement, lacks sufficient development.

\section{Process Enhancement}
\label{sec:9}

As discussed earlier, we have three main elements in process mining, including process discovery, process enhancement, and conformance checking. In the following, initially the reason behind selecting process enhancement as one of the focal point of this paper is discussed. Next, scholarly works is explored and elaborated in relation to specific objectives such as remaining time, drift detection, etc.

In a general sense, when the process mining model is very similar to the model designed for a business or organization, we require a predictive business process monitoring model to monitor and predict bottlenecks, as well as the completion time of the process, etc.~\citep{Bemthuis2019An}. This increases the pace of the improvement process, and the importance of this procedure is determined when process improvement and deployment take longer, causing dissatisfaction at the organizational level~\citep{Volz2016The}. 

On the other hand, process discovery approaches allow for the automated generation of process models from event data~\citep{van2008process}. However, it is common to be interested not just in determining the control flow but also in answering questions about process information (\emph{e.g.}, time remaining). When an organization wants to enhance and re-design its own processes, it must first understand the process characteristics and how they interact with one another. It helps to answer business process management questions for optimizing and improving processes~\citep{gersch2011business}. 

Furthermore, the execution strategy in business operations may evolve over time, resulting in unexpected and unforeseen changes throughout the process. Process modeling itself is incapable of detecting these execution changes, as they often generalize without taking into account new dimensions and assume stable processes~\citep{Ostovar2020}. As a result, focusing on process improvement rather than process modeling is one way to reduce the time and expense of BPM projects. The fundamental benefit of business process predictive monitoring is that it provides information that can be used to take preventive and corrective measures in real time to improve process performance and minimize risks. In other words, this time has already been spent on modeling, and investing it again does not seem very rational. As a result, to compare with current condition, it becomes essential to keep in mind that running instances will be monitored in terms of specific goals~\citep{chan2001applications}. There are four main goals (\emph{i.e.}, cost, time, quality, flexibility) in business process management that could be measured, monitored, and evaluated for this comparison and improvement~~\citep{9241415}.

Upon establishing the desired goals and objectives in process enhancement, the subsequent step is monitoring and evaluating the process. In previous methods, process analysts defined KPIs (Key Performance Indicators) by brainstorming,  benchmarking, and statistical testing \citep{Maaradji2017, Seeliger2017}. However, the acquisition of pertinent data necessitates the collaboration of domain experts~\citep{chandra2021evaluating}.  This approach is not only time-consuming and labor-intensive but also introduces the risk of human errors in judgment during the evaluation process~\citep{issa2016research}.  On the other hand, machine learning and AI have witnessed notable advancements in terms of predictive accuracy, establishing a superiority over alternative techniques with respect to prediction accuracy~\citep{chiorrini2021preliminary}. Moreover, a notable asset of machine learning and AI lies in their capacity to operate without reliance on exhaustive knowledge regarding the underlying process models generating the logs~\citep{baskharon2021predicting}. 

The subsequent sections offer an in-depth analysis of the latest machine learning contributions within the context of process enhancement based on different objectives. 

\subsection{Drift and Anomaly Detection}

Operational processes, especially in agile companies, are constantly undergoing many changes, and as a result ignoring them can cause a drift from the ideal condition and lead to a misalignment with corporate strategies~\citep{Nolle2018}. Similarly, anomalies are characterized by sudden disruptions in processes and demand immediate attention as they can impede process function\citep{Huo2021}.

It is important to note that, drift and anomaly detection are firmly established in quality management practices and concepts, and these notions have witnessed a notable surge in scholarly investigations of business processes in recent academic literature.

With respect to this issue, \citet{Cattafi2010} proposed a solution to consider this in the process of investigating the predictive performance of the process. To address this, they adapt \emph{ IDPML (Inductive Declarative Process Model Learner)} process mining system to create a transition system using annotations. They employed the \emph{ILP (Inductive Logic Programming)}, which combines machine learning and logic programming, to detect changes and identify improvement opportunities as new information becomes available. They believe that reliance on historical process data alone is inefficient. Therefore, they divided the process trace into positive and negative instances. This separation allows them to use the first subset to train a model and establish the normal structure, while the second subset is used to design the process revision system.
On the other hand, ~\citet{Luengo2012} presented an alternative approach to this challenge by applying a clustering method to address the problem. In their article, they offered a method for clustering data that considers the start time of process instances along with other features. The clusters, created by combining control-flow and time characteristics, exhibit both temporal closeness and structural similarity. As a result, the process model developed for each cluster can reflect a unique variation of the business process.

The challenges in gaining insights from the mined models and understanding the findings have not been fully addressed in the health sector. This makes it difficult for health practitioners to apply process mining insights and business process analysis, resulting in the underuse of process mining.
An approach for these areas to enhance their processes is cross-organizational mining and comparative study of processes. A case study by~\citet{Partington2015} presents an examination of how healthcare data and cross-organizational comparisons can be used by process mining tools. They presented findings from a comparative process mining case study using data regularly collected to identify variations in healthcare delivery at four hospitals in Australia.  In another work, in order to reveal behavioral changes,~\citet{Hompes2017} introduced a comparative case clustering approach. Identifying moments in time when behavior changes and understanding the causes of those changes can provide valuable insights and lead to process improvements. Comparing the clustering of two specific event log partitions before and after an identified change point allows for the analysis of behavioral differences. It is possible to identify significant historical moments that can help determine how to divide the event log. This information can then be used for further analysis, such as designing procedures, early identification of undesirable conduct, or meeting auditing requirements.

Data and process analysts utilize visualization as one of the most effective ways to spot and explain changes over time. However, visual study of drift phenomena as systems change over time remains a challenge. This research gap is addressed in~\citep{Yeshchenko2021}. They built a system for fine-granular process drift identification and associated visualizations for performed business process event logs.
They offer several visualizations, such as the Extended Directly- Follows Graph, the Drift Map, Drift Charts, and other measurements to identify the type of drift, for each of these analytical processes.

Utilizing the data logged in event logs to compare the behavior and performance of processes for various process cohorts is a powerful method to obtain evidence-based process improvement insights. Although a number of methods have been put forth to support the various components of such an analysis, the work that has already been done on using event logs to visually compare the performance of various process cohorts still falls short of meeting all the requirements for some analysis scenarios that are of interest to industry. An innovative method that allows visual comparison of such cohorts has been suggested in~\citep{Wynn2017}. Then, a variety of design concepts are suggested to satisfy these needs. The feasibility of creating a functional system based on the suggested principles was proved by a prototype implementation of the technique as part of the process mining framework ProM.

In another work,~\citet{bdcc6040151} employ  customer feedback as a means to identify drift in business processes. They introduce an innovative approach that involves the analysis of real-time feedback to enhance customer satisfaction and evaluate sentiment. The researchers aim to uncover the most important anomalies in the business process based on text mining.

One crucial measure in this context involves identifying potential deviations before they happen, as they could lead to significant costs. Consequently, \citet{10271994} introduced the Business Process Deviation Prediction (BPDP) approach. Their aim was to identify highly influential process characteristics through a two-stage training process (comprising online and offline components) for event classification. This approach empowers process managers to anticipate and prevent deviations during the early stages of running process instances.

Besides drift and anomaly detection in process enhancement, researchers have proposed various methods for predictive monitoring of processes. Each of these methods aims to accurately estimate different parameters. In the following sections, we discuss the various parameters mentioned by the authors.

\subsection{Time Remaining or Completion Time}

Predicting the remaining time of ongoing processes enables and empowers businesses to identify high-risk processes that may not align with certain criteria, thereby equipping case managers with insights for well-informed decisions and timely intervention~\citep{Evermann2017}.~\citet{Aalst2011} offered a customizable method for building a process model, enhancing this model with time data gleaned from prior instances. They employed a set of adaptable abstractions in insurance process cases to produce meaningful temporal prediction by striking a compromise between ``overfitting'' and ``underfitting''. However, these models do not describe how the predicted remaining time will be used.

In reality, these methods use a ``black-box'' strategy since they only predict a single scalar result without breaking it down into simpler parts.

A white-box strategy is suggested in~\citep{Verenich2019Predictingprocessperformance}  to predict performance metrics of  running  process instances. The basic concept is to predict the performance indicator at the level of the activities first, and then, using flow analysis techniques, to aggregate these predictions at the level of a process instance.

To put it another way, the research described here aims to make use of the concept of decompositional explanation. These models help process workers and management understands where the remaining time will be spent in the process. In order to improve the performance and accuracy of remaining-time prediction,  \citet{Wahid2019} suggested a Deep Neural Network (DNN) algorithm in the remaining time prediction issue incorporating categorical variables, which outperforms all baseline approaches used as a benchmark.

\subsection{Next Activity of Running Process}

Another parameter that is in the interest of process analysts is predicting  the next most likely activity of running process instances. Facilitating the process with this valuable and informative insight would enhance operational clarity~\citep{Taymouri2020}. 

There are a number of deep learning architectures that have been extensively studied in machine learning that may be examined as alternatives. They used convolutional neural networks to give the execution scenario of a business process the ability to predict the following action in a running trace.~\citet{Pasquadibisceglie2019} have explored the use of Convolutional Neural Networks (CNNs), which represent a robust class of DNNs widely used in computer vision and speech recognition. The fundamental concept is to consider the temporal data included in the historical event log of a business process as pictures by converting them into spatial data. To do this, each event log trace is first converted into the collection of its prefix traces (\emph{i.e.}\ sequences of events that represent the prefix of a trace). These prefix traces are translated into data structures resembling 2D images. Finally, the created spatial data is utilized to train a Convolutional Neural Network in order to construct a deep learning model capable of predicting the next activity (\emph{i.e.}, the activity associated with the event occurring after the last event in the considered prefix trace). In another study,~\citet{Pasquadibisceglie2020} departed from the approach of using only two attributes (activity and timestamp), as seen in earlier research on predictive process monitoring. Instead, they used the trace representations, similar to the multi-variate time series format to replace the  previous picture format. Consequently, they suggested a picture representation of traces where each pixel captures a feature of the trace scene and is connected with a color value. Therefore, they addressed the challenge of predicting the next activity as a computer vision task and refined this approach in predictive process monitoring research.
In addition to the aforementioned deep learning models, LSTMs (Long Short Term Memory) present a very promising solution for problems related to sequences and time series, such as next activity prediction. However, there is a noticeable gap in the literature regarding the application of such methods.

Decay Replay Mining - Next Transition Prediction (DREAM-NAP) was proposed in~\citep{Theis2019} as a way to enhance the accuracy of the next activity prediction using process state. They add time decay functions to the Petri net model to provide continuous process state samples. Finally, they build a DNN model that predicts the next occurrence using these data together with discrete token movement counters and Petri net marks.  In order to address the challenges associated with limited training data~\citep{Taymouri2020,Kppel2021EvaluatingPB} and uncertainty in event logs~\citep{Prasidis2021}, researchers have introduced an adversarial training framework and Bayesian Networks. This work could benefit a wide range of businesses that struggle with deficiencies and reliability issues in event logs. An alternative promising approach is to use transfer learning, which utilizes information from existing datasets to tailor the model to particular business requirements.

In other work,~\citet{Moon2021} proposed a model called ``POP-O'' (Prediction of Process Using One-Way Language Model Based on NLP Approach) for process prediction. They use a fully attention-based transformer that has shown promising results in recent natural language processing approaches. Companies can adapt to unwelcome changes in their workflow after identifying the name attribute of the event in natural language and predicting the following event. In a comparative research analysis,~\citep{Kratsch2020MachineLI} demonstrate the superior performance of DNN (\emph{i.e.}, simple feedforward deep neural networks and long short-term memory networks) compared to ML techniques (\emph{i.e.}, random forests and support vector machines) in different event logs.

A specific subset of next activity prediction research can be referred to as process outcomes, representing the final status that a process may reach. In a study of road traffic management, an inductive miner was introduced to develop a model capable of predicting these outcomes\citep{10271968}. Additionally, taking into account the outcomes at the task level could further enrich the literature. This method helps organizations save time and effort by enabling task adjustments or prioritizing queues to handle high-risk situations more efficiently.

Next activity prediction is of paramount significance in the context of process enhancement, however, a notable deficiency lies in the lack of comprehensive investigation related to parallel and loop processes. The incorporation of cutting-edge methodologies such as LSTMs, random forest, and recurrent neural networks (RNNs) could be a solid method to address this challenge~\citep{8498218, doganata2017predictive}.

\subsection{Business Constraints Monitoring}

According to the studies, a business constraint is a condition imposed on the execution of a process that distinguishes compliance from non-compliant behavior~\citep{pesic2006declarative}.  In the process execution, decisions are made at various points to satisfy these constraints. As a result, it is critical for process stakeholders to receive predictions on whether the restrictions will be met,along with suggestions on the options that optimize the chances of meeting the business requirements~\citep{poll2018process}.

In this context,~\citet{Maggi2014} present Predictive Business Process Monitoring, a novel framework based on Linear Temporal Logic. This framework continuously generates predictions and recommendations on what activities to perform and what input data values to provide, to determine the likelihood of business constraints being violated. They define a data snapshot as a positive or negative example, depending on whether the constraint was subsequently met in the entire trace. In this method, they translate the prediction problem into a classification challenge, where the aim is to assess whether and with what probability a particular data snapshot leads to business constraint fulfillment. Finally, they use decision tree learning to address the classification issue, producing a decision tree to differentiate between fulfillment and violations. For each potential combination of input attribute values, the decision tree is used to predict the chance that the business constraint will be met. 

Using event log data from activities to make the prediction gives adequate information and decreases the computing cost, whereas prior studies only received a process model with the sequence of events as input, resulting in information loss. To solve this,~\citet{MarquezChamorro2017} suggest an evolutionary decision rule-based system for the prediction of business process indicators. This method's encoding is based on the event properties that were taken from the event logs. A prediction of a certain process indicator is made using the decision rules. Using both next-event and end-of-instance predictions, this system can predict instance-level indicators. 

Another challenge in this context is the substantial runtime expense, which makes it impractical in environments with high throughput or where rapid reaction times are necessary. ~\citet{Francescomarino2019} propose a predictive process monitoring approach for calculating the probability that a particular predicate will be satisfied by the end of a running case. The system considers both the sequence of events in the current trace and the data properties related to these events. By building the classifiers offline, it achieves a reasonably minimal runtime cost. The prediction problem is tackled in two steps. First, prefixes of prior (completed) instances are clustered based on control flow knowledge. Second, a classifier is created for each cluster using the attributes of the event data to differentiate between situations that result in the fulfillment of the predicate and those that result in a violation within the cluster.

\subsection{Performance}

In order to support proactive steps that will enhance the process execution, it should be emphasized that it is more feasible to predict the performance of processes at the level of the process model and identify possible weaknesses in the process~\citep{Park2020}.  In order to achieve this,~\citet{Park2020} present a unique approach to predict future business process performance at the level of the process model. They create performance prediction models based on DNN that take into account the temporal development and spatial dependency of the process model and produce a training set from the process representation matrix.

\subsection{Multi Parameters}

In some papers, the authors focus on process enhancement and consider different objectives. (\emph{e.g.}, remaining time, next activity, etc.). In order to predict sequences of upcoming events, their timestamps, and their associated resource pools, \citet{Camargo2019} suggested a method for training recurrent neural networks (RNN) with Long-Short-Term Memory (LSTM) architecture.  In another work,~\citet{Weinzierl2020} have used Deep Neural Network (DNN) to predict the remaining time and next activity.

In \Cref{Process Enhancement Methods Goals and Objectives}, we categorized papers on process enhancement into different goals: quality, time, cost, and flexibility. These are the main attributes that process analysts pursue in any BPM project. Within these categories, we defined different objectives that the papers focused on and explained them in the corresponding subsections. We also clarified whether each paper included a case study and where it was implemented. From this table, it is evident that flexibility has never been addressed as a goal, indicating a gap in the research. However, this is particularly significant in our fast-paced world with constant changes to the organization.
Most papers have focused on quality, which is also a focal point of AI/ML projects. Having a case study proves to be a crucial factor in evaluating work in this field. Quality emerged as the most frequently pursued goal, with objectives such as drift detection and predicting the next activity receiving significant attention and favorable reviews. Conversely, objectives like detecting bottlenecks and enhancing customer satisfaction require further investigation. We also summarized the AI approaches utilized in the reviewed papers in \Cref{Applied ML-AI Methods in Process Enhancement}. Clustering and neural network methods have been predominantly applied. However, there is a noticeable gap in the implementation of unsupervised learning and semi-supervised techniques, such as Generative Adversarial Networks (GANs), Zero-shot Learning (ZSL), and Variational Autoencoders (VAEs). This gap is particularly significant due to the challenges posed by event logs, which are often difficult to obtain and frequently contain interrupting incomplete traces. This represents a major issue within the field that warrants further exploration.
\begin{landscape}
\begin{table}
\caption{Process enhancement methods goals and objectives}
\label{Process Enhancement Methods Goals and Objectives}
\resizebox{650pt}{!}{%
\begin{tabular}{lccccccccccccc}
\hline
\multicolumn{1}{c}{\multirow{3}{*}{Paper}}       & \multicolumn{4}{c}{Goals}                                                                                     &  & \multicolumn{7}{c}{Objectives}                                                                                                                                                                                                                                                                                                                                                                                                                                                      & \multirow{3}{*}{Case Study}                                                                  \\ \cline{2-5} \cline{7-13}
\multicolumn{1}{c}{}                             & \multirow{2}{*}{Quality}       & \multirow{2}{*}{Time} & \multirow{2}{*}{Cost} & \multirow{2}{*}{Flexibility} &  & \multirow{2}{*}{\begin{tabular}[c]{@{}c@{}}Bottleneck\\ Determination\end{tabular}} & \multirow{2}{*}{\begin{tabular}[c]{@{}c@{}}Customer\\ Satisfaction\end{tabular}} & \multirow{2}{*}{Drift} & \multirow{2}{*}{\begin{tabular}[c]{@{}c@{}}Remaining\\ Time\end{tabular}} & \multirow{2}{*}{\begin{tabular}[c]{@{}c@{}}Next\\ Activity\end{tabular}} & \multirow{2}{*}{\begin{tabular}[c]{@{}c@{}}Business\\ Constraints Monitoring\end{tabular}} & \multirow{2}{*}{Perforemance} &                                                                                              \\
\multicolumn{1}{c}{}                             &                                &                       &                       &                              &  &                                                                                     &                                                                                  &                        &                                                                           &                                                                          &                                                                                            &                               &                                                                                              \\ \cline{1-5} \cline{7-14} 
\citep{Cattafi2010}                              & \checkmark                     &                       &                       &                              &  &                                                                                     &                                                                                  & \checkmark             &                                                                           &                                                                          &                                                                                            & \multicolumn{1}{l}{}          & \begin{tabular}[c]{@{}c@{}}Hotel Management,\\ Auction Management\end{tabular}               \\
\citep{Aalst2011}                                &                                & \checkmark            &                       &                              &  &                                                                                     & \checkmark                                                                       &                        & \checkmark                                                                &                                                                          &                                                                                            & \multicolumn{1}{l}{}          & \begin{tabular}[c]{@{}c@{}}Dutch Insurance\\ Company\end{tabular}                            \\
\citep{Luengo2012}                               &                                &                       &                       &                              &  &                                                                                     &                                                                                  & \checkmark             &                                                                           &                                                                          &                                                                                            & \multicolumn{1}{l}{}          & -                                                                                            \\
\citep{Leoni2014}                                & \checkmark                     &                       &                       &                              &  &                                                                                     & \checkmark                                                                       &                        &                                                                           &                                                                          &                                                                                            & \multicolumn{1}{l}{}          & \begin{tabular}[c]{@{}c@{}}Dutch Employee\\ Insurance Agency\end{tabular}                    \\
\citep{Maggi2014}                                & \checkmark                     &                       &                       &                              &  &                                                                                     &                                                                                  &                        &                                                                           &                                                                          & \checkmark                                                                                 & \multicolumn{1}{l}{}          & -                                                                                            \\
\citep{Partington2015}                           & \checkmark                     &                       & \checkmark            &                              &  & \checkmark                                                                          &                                                                                  & \checkmark             &                                                                           &                                                                          &                                                                                            & \multicolumn{1}{l}{}          & \begin{tabular}[c]{@{}c@{}}4 Australian\\ hospitals\end{tabular}                             \\
\citep{MarquezChamorro2017}                      & \checkmark                     &                       &                       &                              &  &                                                                                     &                                                                                  &                        &                                                                           &                                                                          & \checkmark                                                                                 & \multicolumn{1}{l}{}          & -                                                                                            \\
\citep{Hompes2017}                               & \multicolumn{1}{l}{\checkmark} &                       &                       &                              &  &                                                                                     &                                                                                  & \checkmark             &                                                                           &                                                                          &                                                                                            &                               & \begin{tabular}[c]{@{}c@{}}Dutch academic\\ hospital\end{tabular}                            \\
\citep{Wynn2017}                                 & \multicolumn{1}{l}{\checkmark} &                       & \checkmark            &                              &  & \checkmark                                                                          &                                                                                  & \checkmark             &                                                                           &                                                                          &                                                                                            &                               & \begin{tabular}[c]{@{}c@{}}Insurance\\ Company\end{tabular}                                  \\
\citep{Verenich2019Predictingprocessperformance} & \multicolumn{1}{l}{}           & \checkmark            &                       &                              &  &                                                                                     &                                                                                  &                        & \checkmark                                                                &                                                                          &                                                                                            &                               & -                                                                                            \\
\citep{Wahid2019}                                & \multicolumn{1}{l}{}           & \checkmark            &                       &                              &  &                                                                                     &                                                                                  &                        & \checkmark                                                                &                                                                          &                                                                                            &                               & -                                                                                            \\
\citep{Pasquadibisceglie2019}                    & \multicolumn{1}{l}{\checkmark} &                       &                       &                              &  &                                                                                     &                                                                                  &                        &                                                                           & \checkmark                                                               &                                                                                            &                               &                                                                                              \\
\citep{Theis2019}                                & \multicolumn{1}{l}{\checkmark} &                       &                       &                              &  &                                                                                     &                                                                                  &                        &                                                                           & \checkmark                                                               &                                                                                            &                               & \begin{tabular}[c]{@{}c@{}}9 real-world\\ benchmarks\end{tabular}                            \\
\citep{Francescomarino2019}                      & \checkmark                     &                       &                       &                              &  &                                                                                     &                                                                                  &                        &                                                                           &                                                                          & \checkmark                                                                                 &                               & -                                                                                            \\
\citep{Camargo2019}                              & \checkmark                     & \checkmark            & \checkmark            &                              &  &                                                                                     &                                                                                  &                        & \checkmark                                                                & \checkmark                                                               &                                                                                            & \checkmark                    & -                                                                                            \\
\citep{Pasquadibisceglie2020}                    & \checkmark                     &                       &                       &                              &  &                                                                                     &                                                                                  &                        &                                                                           & \checkmark                                                               &                                                                                            &                               & -                                                                                            \\
\citep{tax2020interdisciplinary}                 & \checkmark                     &                       &                       &                              &  &                                                                                     &                                                                                  &                        &                                                                           & \checkmark                                                               &                                                                                            &                               & \begin{tabular}[c]{@{}c@{}}Compensation\\ Request Process\\ for concert tickets\end{tabular} \\
\citep{Park2020}                                 &                                &                       &                       &                              &  &                                                                                     &                                                                                  &                        &                                                                           &                                                                          &                                                                                            & \checkmark                    &                                                                                              \\
\citep{Kratsch2020MachineLI}                     & \checkmark                     &                       &                       &                              &  &                                                                                     &                                                                                  &                        &                                                                           & \checkmark                                                               &                                                                                            &                               & \begin{tabular}[c]{@{}c@{}}Road Traffic\\ Fine Management\\ Process\end{tabular}             \\
\citep{Yeshchenko2021}                           & \checkmark                     &                       &                       &                              &  &                                                                                     &                                                                                  & \checkmark             &                                                                           &                                                                          &                                                                                            &                               & \begin{tabular}[c]{@{}c@{}}Italian help desk\\ software company\end{tabular}                 \\
\citep{Moon2021}                                 & \checkmark                     &                       &                       &                              &  &                                                                                     &                                                                                  &                        &                                                                           & \checkmark                                                               &                                                                                            &                               & -                                                                                            \\
\citep{bdcc6040151}                              & \checkmark                     &                       &                       &                              &  &                                                                                     &                                                                                  & \checkmark             &                                                                           &                                                                          &                                                                                            &                               & -                                                                                            \\
\citep{10271994}                                 & \checkmark                     &                       & \checkmark            &                              &  &                                                                                     &                                                                                  & \checkmark             &                                                                           &                                                                          &                                                                                            &                               & BPIC/MobIS                                                                                   \\ \hline
\end{tabular}%
}
\end{table}
\end{landscape}

\begin{landscape}
\begin{table}
\caption{Applied ML-AI methods in process enhancement}
\label{Applied ML-AI Methods in Process Enhancement}
\resizebox{640pt}{!}{%
\begin{tabular}{ccccccccccccc}
\hline
\multirow{3}{*}{Paper}                          & \multicolumn{10}{c}{AI Approaches}                                                                                                                                                                                                                                                                                                                                                                                                                                                          & \multirow{3}{*}{Approach Name}                                                         & \multirow{3}{*}{\begin{tabular}[c]{@{}c@{}}Approach\\ Charateristics\end{tabular}} \\ \cline{2-11}
                                                & \multirow{2}{*}{Clustering} & \multirow{2}{*}{Decision Tree} & \multirow{2}{*}{Regression} & \multirow{2}{*}{Classification} & \multirow{2}{*}{XGBoost} & \multicolumn{3}{c}{Neural Network}                                                                                                                                                                              & \multirow{2}{*}{\begin{tabular}[c]{@{}c@{}}Inductive\\ Logic\\ Programming\end{tabular}} & \multirow{2}{*}{NLP} &                                                                                        &                                                                                    \\ \cline{7-9}
                                                &                             &                                &                             &                                 &                          & \begin{tabular}[c]{@{}c@{}}Convolutional \\ Neural Networks\end{tabular} & \begin{tabular}[c]{@{}c@{}}Deep \\ Neural Network\end{tabular} & \begin{tabular}[c]{@{}c@{}}Recurrent \\ Neural Network\end{tabular} &                                                                                          &                      &                                                                                        &                                                                                    \\ \hline
\cite{Cattafi2010}                              &                             &                                &                             &                                 &                          &                                                                          &                                                                &                                                                     & \checkmark                                                                               &                      & \begin{tabular}[c]{@{}c@{}}Inductive Declarative \\ Process Model Learner\end{tabular} & Integritiy                                                                         \\
\cite{Aalst2011}                                &                             &                                &                             &                                 &                          &                                                                          &                                                                &                                                                     &                                                                                          &                      & -                                                                                      & Interpretability                                                                   \\
\cite{Luengo2012}                               & \checkmark                  &                                &                             &                                 &                          &                                                                          &                                                                &                                                                     &                                                                                          &                      & \begin{tabular}[c]{@{}c@{}}Trace Clustering \\ approach\end{tabular}                   & Robustness                                                                         \\
\cite{Leoni2014}                                &                             &                                &                             & \checkmark                      &                          &                                                                          &                                                                &                                                                     &                                                                                          &                      & General Framework                                                                      & Comprehensibility                                                                  \\
\cite{Maggi2014}                                &                             & \checkmark                     &                             &                                 &                          &                                                                          &                                                                &                                                                     &                                                                                          &                      & Linear Temporal Logic                                                                  &                                                                                    \\
\cite{Partington2015}                           & \checkmark                  &                                &                             &                                 &                          &                                                                          &                                                                &                                                                     &                                                                                          &                      & \begin{tabular}[c]{@{}c@{}}Cross-Organizational\\ Mining\end{tabular}                  & Integritiy                                                                         \\
\cite{MarquezChamorro2017}                      &                             &                                &                             &                                 &                          &                                                                          &                                                                &                                                                     &                                                                                          &                      & method's encoding                                                                      & Interpretability                                                                   \\
\cite{Hompes2017}                               & \checkmark                  &                                &                             &                                 &                          &                                                                          &                                                                &                                                                     &                                                                                          &                      & \begin{tabular}[c]{@{}c@{}}Comparative Case\\ Clustering Approach\end{tabular}         &                                                                                    \\
\cite{Wynn2017}                                 & \checkmark                  &                                &                             &                                 &                          &                                                                          &                                                                &                                                                     &                                                                                          &                      & ProcessProfiler3D                                                                      & Interpretability                                                                   \\
\cite{Verenich2019Predictingprocessperformance} &                             &                                &                             & \checkmark                      & \checkmark               &                                                                          &                                                                &                                                                     &                                                                                          &                      & \begin{tabular}[c]{@{}c@{}}Adaptive Flow Analysis\\ White-Box\end{tabular}             & Interpretability                                                                   \\
\cite{Wahid2019}                                &                             &                                &                             &                                 &                          &                                                                          & \checkmark                                                     &                                                                     &                                                                                          &                      & Entity Embedding                                                                       & Integritiy, Accuracy                                                               \\
\cite{Pasquadibisceglie2019}                    &                             &                                &                             &                                 &                          & \checkmark                                                               &                                                                &                                                                     &                                                                                          &                      & -                                                                                      & Robustness                                                                         \\
\cite{Theis2019}                                &                             &                                &                             &                                 &                          &                                                                          & \checkmark                                                     &                                                                     &                                                                                          &                      & DREAM-NAP                                                                              & Agility                                                                            \\
\cite{Francescomarino2019}                      & \checkmark                  &                                &                             &                                 &                          &                                                                          &                                                                &                                                                     &                                                                                          &                      & -                                                                                      & Agility                                                                            \\
\cite{Camargo2019}                              &                             &                                &                             &                                 &                          &                                                                          &                                                                & \checkmark                                                          &                                                                                          &                      & -                                                                                      &                                                                                    \\
\cite{Pasquadibisceglie2020}                    &                             &                                &                             &                                 &                          & \checkmark                                                               &                                                                &                                                                     &                                                                                          &                      & RGB encoding                                                                           & Agility                                                                            \\
\cite{tax2020interdisciplinary}                 &                             &                                &                             &                                 &                          &                                                                          & \checkmark                                                     & \checkmark                                                          &                                                                                          &                      & -                                                                                      & Accuracy                                                                           \\
\cite{Park2020}                                 &                             &                                &                             &                                 &                          &                                                                          & \checkmark                                                     &                                                                     &                                                                                          &                      & DNN                                                                                    &                                                                                    \\
\cite{Kratsch2020MachineLI}                     &                             &                                &                             &                                 &                          &                                                                          & \checkmark                                                     &                                                                     &                                                                                          &                      & -                                                                                      & Accuracy                                                                           \\
\cite{Yeshchenko2021}                           & \checkmark                  &                                &                             &                                 &                          &                                                                          &                                                                &                                                                     & \checkmark                                                                               &                      & \begin{tabular}[c]{@{}c@{}}Visual Drift \\ Detection (VDD)\end{tabular}                & Interpretability                                                                   \\
\cite{Moon2021}                                 &                             &                                &                             &                                 &                          &                                                                          &                                                                &                                                                     &                                                                                          & \checkmark           & POP-ON                                                                                 & Interpretability                                                                   \\
\cite{bdcc6040151}                              & \checkmark                  &                                &                             & \checkmark                      &                          &                                                                          &                                                                &                                                                     &                                                                                          &                      & \begin{tabular}[c]{@{}c@{}}Lexicon-based\\ Sentiment Analysis\end{tabular}             & Agility, Comprehensibility                                                         \\
\citep{10271994}                                &                             &                                &                             & \checkmark                      &                          &                                                                          &                                                                &                                                                     &                                                                                          &                      & \begin{tabular}[c]{@{}c@{}}Two-stage Offline/Online\\  Classification\end{tabular}     & Accuracy                                                                           \\ \hline
\end{tabular}%
}
\end{table}
\end{landscape}

\section{Process Improvement}

As discussed earlier, process improvement is used to optimize time, cost, quality, and flexibility. It includes general methods including six sigma, company restructuring, core process redesign, and continuous improvement~\citep{aladwani2001change}. It might be difficult for process analysts to decide which business analysis methods and techniques are best for identifying issues and guiding the choice of suitable analysis approaches~\citep{Zellner2013}.

As mentioned previously, event logs are mined for non-trivial and practical information using process enhancement techniques. The goal of process improvement, on the other hand, is to find a solution to the problem of how to redesign the process so that business operations are effective and efficient. Understanding all components of the existing business process situation that are relevant to future process improvement is the overarching goal of process enhancement~\citep{Khosravi2016}. Recent advances in research have made the fusion of these two approaches one of the most frequently employed areas of process analysis in enterprises, businesses, and governments~\citep{Song2011}. 

According to the framework provided in~\citep{grant2002wider},  the business redesign includes five critical dimensions: (1) process/task, (2) technology, (3) people, (3) communication, and (4) organizational structure, which interact dynamically within the business. Each of these holds the potential for business improvement.  Among them, within the process dimension, there are essential elements such as:
\begin{enumerate}
    \item  Activity Definition~\citep{jablonski1996workflow} (identifying necessary tasks or steps to achieve the process goal, naming, and determining required actions in each step),
    \item Resource Allocation and Scheduling~\citep{alter1991information} (selecting the most suitable resource for each step, and defining appropriate timing for tasks execution),
    \item  Process Objectives~\citep{butler1996strategic} (determining the desired output upon process completion),
    \item  Process Trigger~\citep{hammer2009reengineering} (specific events that initiate the execution of a process (\emph{e.g.}, customer request or bug in software)),
    \item Precedence and Dependency ~\citep{klein199510} (the sequence and interdependence of activities, defining the activity order),
\end{enumerate}

that serve as potential focal points for researchers engaged in BPM projects~\citep{seidmann1997effects}. 

~\citet{Grant2016} examined how these techniques for business process improvement might be used to help the creation of effective and efficient business processes. The author contends that despite the availability of several approaches and procedures to enable business analysis and redesigning, there is a dearth of thorough instructions on how to use these techniques in actual practice. He introduced the current techniques including (1) Root cause analysis (RCA)~\citep{arnheiter2008looking, ming2007root},  (2) Business process analysis and activity elimination (BP\&AE)~\citep{}, (3) Duration analysis (DA)~\citep{fooladi2000risk}, (4) Problem analysis (PA)~\citep{dennis2015systems}, (5) Activity-based costing (ABC)~\citep{brierley2011proper}, (6) Benchmarking (BM)~\citep{anand2008benchmarking}, (7) Outcome analysis (OA), (8) Technology analysis (TA)~\citep{caldeira2010we}.

Hence, the process improvement evolution can be outlined into three distinct stages: 
First, the traditional approaches primarily employ statistical methods such as Lean, Six Sigma as well as benchmarking. Subsequently,  the second stage involves the adoption of heuristic approaches which are discussed thoroughly in ~\citep{reijers2005best}. This stage predominately emphasizes continuous improvement strategies and the utilization of brainstorming techniques.
Lastly, the third stage includes AI approaches which have emerged due to the advancement in big data and machine learning methods, forming a relatively new area that requires further elucidation.
Consequently, there are a lot of research gaps in this method because of the numerous uses and early age of the method. In the following, we will discuss these methods and research gaps in detail.

\subsection{Pattern Similarity}

Pattern recognition and feature extraction are established concepts in the domain of machine learning~\citep{tolciu2021analysispatterns}. In the context of business process management, pattern similarity plays a crucial role in the ongoing evaluation and comparison of a process against predefined benchmarks. This assessment occurs during the execution of the process, ensuring its efficiency can be measured reliably. Moreover, the process can be iteratively redesigned with targeted elements to achieve the ultimate objective~\citep{SOLA2022102049}. Indeed, pattern similarity can be acknowledged as a significant contribution of AI to the practice of benchmarking~\citep{shafagatova2021alignment}.

\citet{Niedermann2010} present an innovative approach for optimizing business processes at the design stage, rather than waiting until the implementation stage. The authors provide a comprehensive review of related work in the field, highlighting the limitations of existing approaches and the need for a more comprehensive solution. The proposed approach leverages machine learning techniques to identify optimization patterns in process models and match these patterns with similar patterns in a database of known patterns, providing a framework that consists of three main steps: identifying optimization patterns, matching process models, and applying optimization patterns. This paper contributes to the field of business process optimization by proposing a novel approach to optimize business processes at the design stage that can lead to significant improvements in performance and reduce the time and cost of process optimization.

Making informed decisions regarding the implementation of organizational changes or process-aware information systems (PAISs) requires a priori assessment of process improvement patterns (PIPs) for a particular application scenario and consideration of an organization's strategy, goals, existing business processes, and information systems landscape. According to~\citep{Lohrmann2016}, there is a gap in the literature on the absence of precise instructions on how to successfully implement process improvement patterns (PIP) in the context of business processes. The authors provide a method that entails applying suitable patterns for process improvement, translating those patterns to the particular business process setting, and then implementing the patterns to the process. This entails choosing the most suitable patterns to use for the process by comparing the features and needs of the process with the patterns in the repository. The procedure may then be improved, making it more productive and efficient.

Conventional techniques like audits, walkthroughs, and simulations can be time-consuming, and expensive, and they might not catch all potential flaws ~\citep{Bergener2015}. Using semantic pattern matching in process models,~\citet{Bergener2015} offered a novel method for locating probable drawbacks in business processes. The authors suggest a system that evaluates process models automatically and finds semantic patterns that might point to potential flaws. The method enables businesses to identify potential flaws in their operational procedures quickly and precisely, which improves decision-making and operational procedures. The authors' proposed methodology consists of three main steps: (1) preprocessing process models to extract key features; (2) identifying semantic patterns by clustering key features with a graph-based algorithm; and (3) evaluating semantic patterns to ascertain their potential as indicators of weaknesses in the process model.

\subsection{Keyword Extraction}

The present manual methods of business process redesign have the drawbacks of being time-consuming, subjective, and prone to human error~\citep{Mustansir2022}.
In other work, by automating the improvements,~\citet{Mustansir2022} want to increase the efficacy and efficiency of the redesign process. The NLP-based method for automatically extracting redesign recommendations for business processes is proposed in this research. The suggested method identifies sections of a business process that may be improved by combining many approaches, such as keyword extraction, dependency parsing, and semantic similarity metrics. This NLP-based method extracts suggestions for redesign from textual descriptions of business operations. The method's objective is to find process areas that can be improved, such as omitting redundant steps, streamlining communication, or cutting inefficiencies.

In knowledge-intensive processes characterized by high uncertainty in the execution of running processes, there exist multiple choices to transition a task to another activity. However, it is important to find the best suitable trace for the case to make it efficient and prevent waste or redundancy. In this way, \citet{KHANDAKER2024107450} suggested using a pre-trained transformer to bridge the gap between textual information and process semantics on little or even no labeled email data, aiming to find and recommend the best candidate for the next activity. In a similar work, a Generative Pre-trained Transformer (ProcessGPT) model on a large dataset of business process data has been introduced \citep{10248283} to not only find an efficient process model but also generate new process models to adapt to customer needs and enhance process automation within the organization based on context and user input.

\subsection{Lean Six Sigma}

The utilization of machine learning techniques in process redesign is subject to limitations, which consequently impact the effectiveness of process improvement~\citep{Al-Anqoudi2021}. Given that all methodologies must adhere to expert recommendations and be executed manually, their efficiency within enterprise organizations is compromised~\citep{Hashem2019OrganizationalEO}. In an effort to automate traditional methods of business process re-engineering,~\citet{9850932} pursued the application of Lean Six Sigma principles and waste identification through labelled data. Consequently, they proposed an automated process that involves the classification of actions, identification of waste, and the subsequent removal and merging of activities within the process flow.

\subsection{Resource Scheduling}

Due to the noticeable costs associated with human resources, efficient resource scheduling holds significant importance, and multiple approaches are delivered in BPM including multi-criteria resource allocation~\citep{Arias2018HumanRA}, ability-based human resource allocation~\citep{Erasmus2018AMT}, etc. While machine learning methods have emerged as a dynamic area of research, their application in business process management, specifically in predicting resource demand (\emph{e.g.}, the number of replicas) based on partial execution, remains limited \citep{mehdiyev2020novel, pfeiffer2021multivariate}.
However, the application of deep learning in resource scheduling has witnessed considerable advancements in diverse domains, including production and project management, various approaches, such as rewriting logic~\citep{duran2019rewriting}, support vector machine~\citep{DENKENA2016221}, etc., have been introduced to tackle resource scheduling in business domain~\citep{wilmott2019machine}. Reinforcement learning is an alternative approach that has been explored in the domain of  business process cycle-time optimization. \citet{Firouzian2019} adopted an entropy-based learning approach, aimed at increasing task similarity during task assignment. The underlying reason contends that cycle time has a negative correlation with assigning similar tasks to the same resource. To fill the gap, an approach has been delivered, focusing on the analysis of business process provisioning under a resource prediction strategy based on LSTM~\citep{Duran2022}.  The proposed approach takes as input a BPMN process and a set of traces to train an LSTM model. By utilizing these inputs, the model predicts the optimal allocation and release of resources, aiming to optimize their utilization. This optimization can involve minimizing the idle time of resources or maximizing their usage across the business process while associated with the execution of the process. In addition, machine learning holds promise in automating resource assignment~\citep{ALHAWARI2021702, Ahmed2022MultipleSC,Agarwal2020AutomatedAO, khadivi2023deep}  and routing problems \citep{Mandal2019ImprovingIS,Nikulin2021ApplicationOM,Paramesh2018AutomatedIS,Gamboa2022FurtherEO}, particularly within support processes. they proposed the implementation of machine learning models for ticket classification, enabling the proper association with their respective services. They have used ticket information including description, comments, attachments, and other pertinent details, to establish an automated process improvement while simultaneously attending to customer needs in a dynamic manner.

\subsection{Sustainability}

The lack of sustainable business process management (BPM) models for the construction industry was a gap in the literature. This strategy is suggested by ~\citet{Song2011}  to address the requirement for a sustainable BPM model that takes into account social, economic, and environmental aspects. The sustainable BPM model for construction businesses is the contribution. Strategic planning, process design, and process implementation and monitoring make up the model's three stages. The concept aims to raise social responsibility among construction enterprises, lower their environmental impact, and boost their financial success.

In summary, based on the scope of process redesign, papers are categorized, and the implemented case studies are highlighted in \Cref{Applied ML-AI Approaches in Process Improvement  Methods }.

\begin{table}[h]

\centering
\caption{Process redesign scope in process improvement}
\label{Applied ML-AI Approaches in Process Improvement  Methods }
\resizebox{\columnwidth}{!}{%
\begin{tabular}{ccccccccccccccccccccccccccccccc}
\cline{1-7}
\hline
\multirow{3}{*}{Paper}           & \multicolumn{5}{c}{Process Redesign Scope}                                                                                                                                                                                                                                                                                                                     & \multirow{3}{*}{Case Study}                           \\ \cline{2-6}
                                 & \multirow{2}{*}{\begin{tabular}[c]{@{}c@{}}Activity\\ Definition\end{tabular}} & \multirow{2}{*}{\begin{tabular}[c]{@{}c@{}}Resource\\ Assign\end{tabular}} & \multirow{2}{*}{\begin{tabular}[c]{@{}c@{}}Process\\ Objectives\end{tabular}} & \multirow{2}{*}{\begin{tabular}[c]{@{}c@{}}Process\\ Trigger\end{tabular}} & \multirow{2}{*}{\begin{tabular}[c]{@{}c@{}}Precedence\\ and Dependency\end{tabular}} &                                                       \\
                                 &                                                                                &                                                                            &                                                                               &                          &                                                                                      &                                                       \\ \hline
\citep{Niedermann2010}           & \checkmark                                                                     &                                                                            & \checkmark                                                                    &                          &                                                                                      & \begin{tabular}[c]{@{}c@{}}Retail\\ Bank\end{tabular} \\
\citep{Bergener2015}             & \checkmark                                                                     &                                                                            &                                                                               &                          &                                                                                      & Bank                                                  \\
\citep{Lohrmann2016}             & \checkmark                                                                     &                                                                            &                                                                               &                          &                                                                                      & HR Management                                         \\
\citep{Paramesh2018AutomatedIS}  &                                                                                &                                                                            &                                                                               &                          & \checkmark                                                                           & IT Help Desk                                          \\
\citep{Mandal2019ImprovingIS}    &                                                                                &                                                                            &                                                                               &                          & \checkmark                                                                           & IT Help Desk                                          \\
\citep{Firouzian2019}            &                                                                                & \checkmark                                                                 &                                                                               &                          &                                                                                      & BPI 2012                                              \\
\citep{ALHAWARI2021702}          &                                                                                & \checkmark                                                                 &                                                                               &                          &                                                                                      & IT Help Desk                                          \\
\citep{Nikulin2021ApplicationOM} &                                                                                &                                                                            &                                                                               &                          & \checkmark                                                                           & IT Help Desk                                          \\
\citep{Mustansir2022}            & \checkmark                                                                     &                                                                            & \checkmark                                                                    &                          &                                                                                      & -                                                     \\
\citep{Al-Anqoudi2021}           & \checkmark                                                                     & \checkmark                                                                 & \checkmark                                                                    &                          &                                                                                      & -                                                     \\
\citep{Duran2022}                &                                                                                & \checkmark                                                                 &                                                                               &                          & \checkmark                                                                           & -                                                     \\
\citep{Ahmed2022MultipleSC}      &                                                                                &                                                                            &                                                                               &                          & \checkmark                                                                           & -                                                     \\
\citep{10248283}                 & \checkmark                                                                     &                                                                            &                                                                               &                          & \checkmark                                                                           & -                                                     \\
\citep{KHANDAKER2024107450}      & \checkmark                                                                     &                                                                            &                                                                               &                          & \checkmark                                                                           & -                                                     \\ \hline
\end{tabular}%
}
\end{table}

\begin{landscape}
\begin{table}

\caption{Process improvement methods summary}
\label{Process Improvement Methods Summary}
\resizebox{650pt}{!}{%
\begin{tabular}{lccccccccccccccclccc}
\hline
\multicolumn{1}{c}{\multirow{3}{*}{Paper}} & \multicolumn{5}{c}{Method}                                                                                                                                                                                                                                                                                                                                                                                     &                      & \multicolumn{11}{c}{AI Approaches}                                                                                                                                                                                                                                                                                                                                                                                                                                                                                                                 & \multirow{3}{*}{\begin{tabular}[c]{@{}c@{}}Approach\\ Name\end{tabular}}                   & \multirow{3}{*}{\begin{tabular}[c]{@{}c@{}}Approach\\ Characteristics\end{tabular}} \\ \cline{2-6} \cline{8-18}
\multicolumn{1}{c}{}                       & \multirow{2}{*}{\begin{tabular}[c]{@{}c@{}}Pattern\\ Similarity\end{tabular}} & \multirow{2}{*}{\begin{tabular}[c]{@{}c@{}}Keyword \\ Extraction\end{tabular}} & \multirow{2}{*}{\begin{tabular}[c]{@{}c@{}}Lean \\ Six\\ Sigma\end{tabular}} & \multirow{2}{*}{\begin{tabular}[c]{@{}c@{}}Resource\\ Scheduling\end{tabular}} & \multirow{2}{*}{\begin{tabular}[c]{@{}c@{}}Dependency\\ Parsing\end{tabular}} &                      & \multirow{2}{*}{Clustering} & \multirow{2}{*}{SVM} & \multirow{2}{*}{Classification} & \multirow{2}{*}{XGBoost} & \multicolumn{3}{c}{Neural Network}                                                                                                                                & \multirow{2}{*}{\begin{tabular}[c]{@{}c@{}}Reinforcement\\ Learning\end{tabular}} & \multicolumn{1}{l}{\multirow{2}{*}{\begin{tabular}[c]{@{}l@{}}Generative\\ Adversarial\\ Networks\end{tabular}}} & \multirow{2}{*}{Transformer}   & \multirow{2}{*}{NLP} &                                                                                            &                                                                                     \\ \cline{12-14}
\multicolumn{1}{c}{}                       &                                                                               &                                                                                &                                                                              &                                                                                &                                                                               &                      &                             &                      &                                 &                          & LSTM                 & \begin{tabular}[c]{@{}c@{}}Deep\\ Neural\\ Network\end{tabular} & \begin{tabular}[c]{@{}c@{}}Convolutional\\ Neural\\ Network\end{tabular} &                                                                                   & \multicolumn{1}{l}{}                                                                                             &                                &                      &                                                                                            &                                                                                     \\ \hline
\citep{Niedermann2010}                     & \checkmark                                                                    &                                                                                &                                                                              &                                                                                &                                                                               &                      &                             &                      &                                 &                          &                      & \checkmark                                                      &                                                                          &                                                                                   &                                                                                                                  &                                &                      & \begin{tabular}[c]{@{}c@{}}deep\\ Business\\ Optimization\\ Platform\\ (dBOP)\end{tabular} & Agility                                                                             \\
\citep{Bergener2015}                       & \checkmark                                                                    &                                                                                &                                                                              &                                                                                &                                                                               &                      & \checkmark                  &                      &                                 &                          &                      &                                                                 &                                                                          &                                                                                   &                                                                                                                  &                                &                      & \begin{tabular}[c]{@{}c@{}}Semantic\\ Pattern\\ Matching\end{tabular}                      & \begin{tabular}[c]{@{}c@{}}Agility\\ Interpretability\end{tabular}                  \\
\citep{Lohrmann2016}                       & \checkmark                                                                    &                                                                                &                                                                              &                                                                                &                                                                               &                      &                             &                      &                                 &                          &                      & \checkmark                                                      &                                                                          &                                                                                   &                                                                                                                  &                                &                      & PIP                                                                                        & Agility                                                                             \\
\citep{Paramesh2018AutomatedIS}            &                                                                               &                                                                                &                                                                              &                                                                                &                                                                               &                      &                             & \checkmark           &                                 &                          &                      &                                                                 &                                                                          &                                                                                   &                                                                                                                  &                                &                      & -                                                                                          & Accuracy                                                                            \\
\citep{Mandal2019ImprovingIS}              &                                                                               &                                                                                &                                                                              &                                                                                &                                                                               &                      &                             &                      & \checkmark                      &                          &                      &                                                                 & \checkmark                                                               &                                                                                   &                                                                                                                  &                                & \checkmark           & \begin{tabular}[c]{@{}c@{}}Multi-Modal\\ Analysis\end{tabular}                             & -                                                                                   \\
\citep{Firouzian2019}                      &                                                                               &                                                                                &                                                                              & \checkmark                                                                     &                                                                               &                      &                             &                      &                                 &                          &                      &                                                                 &                                                                          & \checkmark                                                                        &                                                                                                                  &                                &                      & \begin{tabular}[c]{@{}c@{}}Entropy-Based\\ Learning\end{tabular}                           & Comprehensibility                                                                   \\
\citep{ALHAWARI2021702}                    &                                                                               &                                                                                &                                                                              & \checkmark                                                                     &                                                                               &                      &                             &                      & \checkmark                      &                          &                      &                                                                 &                                                                          &                                                                                   &                                                                                                                  &                                &                      & -                                                                                          & -                                                                                   \\
\citep{Nikulin2021ApplicationOM}           &                                                                               &                                                                                &                                                                              & \checkmark                                                                     &                                                                               &                      &                             &                      & \checkmark                      &                          &                      &                                                                 &                                                                          &                                                                                   &                                                                                                                  &                                &                      & -                                                                                          & Agility                                                                             \\
\citep{Mustansir2022}                      & \checkmark                                                                    & \checkmark                                                                     &                                                                              &                                                                                & \multicolumn{2}{c}{\checkmark}                                                                       &                             &                      &                                 &                          &                      & \checkmark                                                      &                                                                          &                                                                                   &                                                                                                                  &                                & \checkmark           & -                                                                                          & -                                                                                   \\
\citep{Al-Anqoudi2021}                     &                                                                               &                                                                                & \checkmark                                                                   &                                                                                &                                                                               &                      &                             &                      &                                 & \checkmark               &                      &                                                                 &                                                                          &                                                                                   &                                                                                                                  &                                &                      & -                                                                                          & -                                                                                   \\
\citep{Duran2022}                          &                                                                               &                                                                                &                                                                              & \checkmark                                                                     &                                                                               &                      &                             &                      &                                 &                          &                      &                                                                 &                                                                          &                                                                                   &                                                                                                                  &                                &                      & -                                                                                          & -                                                                                   \\
\citep{Ahmed2022MultipleSC}                &                                                                               &                                                                                &                                                                              & \checkmark                                                                     &                                                                               &                      &                             &                      & \checkmark                      &                          & \checkmark           &                                                                 &                                                                          &                                                                                   &                                                                                                                  &                                &                      & \begin{tabular}[c]{@{}c@{}}Rewriting\\ Logic\\ Semantics\end{tabular}                      & Agility                                                                             \\
\citep{10248283}                           & \multicolumn{1}{l}{}                                                          & \checkmark                                                                     & \multicolumn{1}{l}{}                                                         & \multicolumn{1}{l}{}                                                           & \multicolumn{1}{l}{}                                                          & \multicolumn{1}{l}{} & \multicolumn{1}{l}{}        & \multicolumn{1}{l}{} & \multicolumn{1}{l}{}            & \multicolumn{1}{l}{}     & \multicolumn{1}{l}{} & \multicolumn{1}{l}{}                                            & \multicolumn{1}{l}{}                                                     & \multicolumn{1}{l}{}                                                              & \checkmark                                                                                                       &                                & \checkmark           & ProcessGPT                                                                                 & Robustness                                                                          \\
\citep{KHANDAKER2024107450}                & \multicolumn{1}{l}{}                                                          & \checkmark                                                                     & \multicolumn{1}{l}{}                                                         & \multicolumn{1}{l}{}                                                           & \multicolumn{1}{l}{}                                                          & \multicolumn{1}{l}{} & \multicolumn{1}{l}{}        & \multicolumn{1}{l}{} & \multicolumn{1}{l}{}            & \multicolumn{1}{l}{}     & \multicolumn{1}{l}{} & \multicolumn{1}{l}{}                                            & \multicolumn{1}{l}{}                                                     & \multicolumn{1}{l}{}                                                              &                                                                                                                  & \multicolumn{1}{c}{\checkmark} & \checkmark           & Task Mapper                                                                                & -                                                                                   \\ \hline
\end{tabular}%
}
\end{table}
\end{landscape}

The concept of process triggers is identified as a gap in the existing literature and presents a potential direction for future research in process improvement methods. Conversely, the definition of activities within processes is a well-explored area, with numerous papers over the past decade demonstrating its potential and continued relevance for further study.
Selecting the appropriate scope for process improvement is crucial, as it depends on various factors such as process characteristics, stakeholders involved, and other pertinent considerations.

In addition to the scope, in \Cref{Process Improvement Methods Summary}, we discussed the different methods and AI approaches found in the literature that can be beneficial for process improvement. Various approaches have been explored in the literature, demonstrating a diverse range of methodologies. However, some approaches, such as clustering, Reinforcement Learning (RL), Generative Adversarial Networks (GANs), and Transformer models, have received comparatively less attention in research. Furthermore, the characteristics of each approach were specified based on the emphasis of the paper, including agility, interpretability, comprehensibility, and robustness, each of which is essential for the proposed approaches.
\section{Process Enhancement and Process Improvement}
\noindent

Event logs are mined for non-trivial and practical information using process enhancement techniques. The goal of process improvement, on the other hand, is to find a solution to the problem of how to redesign the process so that business operations are effective and efficient. Understanding all components of the existing business process situation that are relevant to future process improvement is the overarching goal of process enhancement~\citep{VanDerAalst2015}.
Recent advances in research have made the fusion of these two approaches one of the most frequently employed areas of process analysis in enterprises, businesses, and governments. Consequently, there are a lot of research gaps in this method because of the numerous uses and early age of the method. In the following, we will discuss these research gaps in detail.

\subsection{Simulation}
A few simulation studies additionally take into account Process Mining as a support tool to more accurately depict the behavior of the agents.  Event data analysis may show a variety of performance and compliance issues as well as provide suggestions for performance enhancements. A unique method for constructing system dynamics models for simulation in the context of operational processes was put out in~\citep{Pourbafrani2020}.

Their method allows for the detection and aggregate modeling of the effects and relations at the instance level. They give an interactive platform for modeling the performance metrics as system dynamics models and extract a number of performance factors from the process's present state using past execution data. The concept of combining system dynamics with process mining at the aggregate level is introduced, which results in the building of models that take external influences into account. Process enhancement tools are well-supported in identifying ongoing processes and deviations as well as identifying performance issues and bottlenecks. All of the strategies offered, however, reflect the process's historical and contemporary states. Understanding the potential actions and modifications that may be taken is necessary for process improvement. Using an interface, ~\citet{Pourbafrani2021} introduce a novel tool that enables process owners to automatically extract all the process characteristics from their historical event data, make changes to these aspects, and then re-run the process. Process mining and simulation techniques together provide fresh, evidence-based approaches to ponder hypothetical scenarios. This tool helps to evaluate performance factors including resource, activity, process capacity, and time while implementing efficient modifications to the process. Both process improvement and interactive process enhancement are directly addressed by these capabilities. 

On the other hand, in some works~\citep{ANBALAGAN20217135, CZVETKO2022117} directed their attention toward a manufacturing process, which they approached from a business process perspective to evaluate and improve. They adopted an integrated Industry 4.0 and business management approach to monitor and constantly  develop business processes. To facilitate process redesign, they automatically map the (Business Process Modeling and Notations) BPMN model into simulation code that is prepared for execution, after which they annotate the model with the essential information to make it executable. ~\citet{Suriadi2017} suggest a method for examining one element of resource behavior: the way a resource prioritizes his or her workload. The method uses transactional data from event logs to learn how to prioritize resources when carrying out work. The method has been expanded to include teaching the instances and activities' priority rankings.

The fusion of DNN and discrete event simulation is another area of research focus. For instance, \citet{10271993} directed their efforts towards enhancing business process performance by incorporating queue dynamics as an intercase feature in the DNN models. The RIMS (Runtime Integration of Machine Learning and Simulation) method facilitates a seamless integration of cycle time predictions for each trace within a DNN model during runtime simulation.

\subsection{Root-Cause Analysis}
Results from the process improvement can highlight problems with the organization's process management. Half of a problem is solved by clearly stating and clarifying it. In other words, process improvement strategies can effectively and efficiently identify solutions for them ~\citep{spencer1994models}. 
For this, ~\citet{Lehto2016} introduced influence analysis, a general approach for locating business improvement opportunities connected to business processes. Process mining, root cause analysis, and classification rule mining are the foundations of influence analysis. They go on to specify metrics for informing business people about the contribution results and demonstrate how these metrics might be employed to concentrate on advancements. Decisions on whether to modify the entire process design or just a few sticking points require consideration of influence analysis. Behaviors that are not reflected in the process model derived by the process mining method may lead to process improvement. \citet {Dees2017} presents an approach for enhancing a process model to include observable behavior. The model-repair approach is paired with classification-tree learning techniques to decide which deviations to include in the revised model. The classification tree analyzes if and to what extent KPI values and observed behavior are correlated. The model is then improved by including behavior that was not considered in the original model but is associated with higher KPI levels while not contravening corporate policies.
Organizations can identify unwanted working patterns and investigate contextual elements that can unintentionally support the manifestation of these behaviors by workers with the help of a clear knowledge of resource behavior, providing a clear path for process development. There are few methods for understanding resource behavior, despite the fact that several process mining approaches have been presented to extract insights about how activities inside processes are carried out. The management of resources must be clever and effective in order to permit this key aspect, which has a substantial impact on the performance of the entire process. 

In \Cref{Combined Process Enhancement and Process Improvement Methods Summary}, we showed all the papers that addressed both process enhancement and process improvement. 

\begin{sidewaystable*}[htbp]

\caption{Combined process enhancement and process improvement methods summary}
\label{Combined Process Enhancement and Process Improvement Methods Summary}
\resizebox{670pt}{!}{%
\begin{tabular}{ccccccccccccccccclcc}
\hline
\multirow{3}{*}{Paper}    & \multicolumn{4}{c}{Goal}                                                                                &  & \multicolumn{7}{c}{Method}                                                                                                                                                                                                                                                                                                                                                                                                                                                                                                      &  & \multicolumn{4}{c}{AI Approaches}                                                                                                                                         & \multirow{3}{*}{\begin{tabular}[c]{@{}c@{}}Approach\\ Name\end{tabular}}                            & \multirow{3}{*}{Case Study}                                                             \\ \cline{2-5} \cline{7-13} \cline{15-18}
                          & \multirow{2}{*}{Quality} & \multirow{2}{*}{Time} & \multirow{2}{*}{Cost} & \multirow{2}{*}{Flexibility} &  & \multirow{2}{*}{\begin{tabular}[c]{@{}c@{}}Activity\\ Definition\end{tabular}} & \multirow{2}{*}{\begin{tabular}[c]{@{}c@{}}Resource\\ Assign\end{tabular}} & \multirow{2}{*}{\begin{tabular}[c]{@{}c@{}}Process\\ Objectives\end{tabular}} & \multirow{2}{*}{\begin{tabular}[c]{@{}c@{}}Process\\ Trigger\end{tabular}} & \multirow{2}{*}{\begin{tabular}[c]{@{}c@{}}Precedence and \\ Dependency\end{tabular}} & \multirow{2}{*}{Simulation} & \multirow{2}{*}{\begin{tabular}[c]{@{}c@{}}Root-Cause\\ Analysis\end{tabular}} &  & \multirow{2}{*}{Clustering} & \multirow{2}{*}{\begin{tabular}[c]{@{}c@{}}Decision\\ Tree\end{tabular}} & \multirow{2}{*}{Classification} & \multirow{2}{*}{DNN} &                                                                                                     &                                                                                         \\
                          &                          &                       &                       &                              &  &                                                                                &                                                                            &                                                                               &                                                                            &                                                                                       &                             &                                                                                &  &                             &                                                                          &                                 &                                &                                                                                                     &                                                                                         \\ \cline{1-5} \cline{7-13} \cline{15-20} 
\citep{Lehto2016}         & \checkmark               &                       &                       &                              &  & \checkmark                                                                     &                                                                            & \checkmark                                                                    &                                                                            &                                                                                       &                             & \checkmark                                                                     &  &                             &                                                                          & \checkmark                      &                                & \begin{tabular}[c]{@{}c@{}}Influence\\ Analysis\end{tabular}                                        & \begin{tabular}[c]{@{}c@{}}Rabobank\\ Group ICT\end{tabular}                            \\
\citep{Dees2017}          & \checkmark               &                       &                       &                              &  & \checkmark                                                                     &                                                                            & \checkmark                                                                    &                                                                            & \checkmark                                                                            &                             & \checkmark                                                                     &  &                             & \checkmark                                                               & \checkmark                      &                                & \begin{tabular}[c]{@{}c@{}}Model-Repair\\ Technique\end{tabular}                                    & UWV                                                                                     \\
\citep{Suriadi2017}       &                          &                       & \checkmark            &                              &  &                                                                                & \checkmark                                                                 &                                                                               &                                                                            &                                                                                       &                             & \checkmark                                                                     &  & \checkmark                  &                                                                          &                                 &                                & -                                                                                                   & \begin{tabular}[c]{@{}c@{}}Australian\\ Insurance\\ Company\end{tabular}                \\
\citep{Pourbafrani2020}   &                          & \checkmark            &                       &                              &  & \checkmark                                                                     &                                                                            &                                                                               &                                                                            &                                                                                       & \checkmark                  &                                                                                &  & \checkmark                  &                                                                          &                                 &                                & \begin{tabular}[c]{@{}c@{}}Scenario-based \\ predictions\end{tabular}                               & -                                                                                       \\
\citep{Pourbafrani2021}   & \checkmark               & \checkmark            & \checkmark            &                              &  & \checkmark                                                                     & \checkmark                                                                 &                                                                               &                                                                            &                                                                                       & \checkmark                  &                                                                                &  & \checkmark                  &                                                                          &                                 &                                & \begin{tabular}[c]{@{}c@{}}Semantic\\ Pattern\\ Matching\end{tabular}                               & -                                                                                       \\
\citep{ANBALAGAN20217135} &                          & \checkmark            &                       &                              &  & \checkmark                                                                     &                                                                            &                                                                               &                                                                            & \checkmark                                                                            &                             &                                                                                &  &                             &                                                                          &                                 &                                & -                                                                                                   & -                                                                                       \\
\citep{Czvetko2022}       & \checkmark               &                       &                       &                              &  & \checkmark                                                                     &                                                                            &                                                                               &                                                                            &                                                                                       & \checkmark                  &                                                                                &  &                             &                                                                          & \checkmark                      &                                & I4.0 -CRISP-ML                                                                                      & \begin{tabular}[c]{@{}c@{}}Hungarian assembly \\ and\\ engineering company\end{tabular} \\
\citep{10271993}          &                          & \checkmark            & \checkmark            &                              &  &                                                                                & \checkmark                                                                 &                                                                               &                                                                            &                                                                                       & \checkmark                  &                                                                                &  &                             &                                                                          &                                 & \multicolumn{1}{c}{\checkmark} & \begin{tabular}[c]{@{}c@{}}Runtime Integration\\ of Machine Learning \\ and Simulation\end{tabular} & BPIC                                                                                    \\ \hline
\end{tabular}%
}

\end{sidewaystable*}

Similar to process enhancement, flexibility remains an unexplored goal in the context of combined process enhancement and process improvement. Most of the methods referred to redefining activities within the process. However, the topic of improving process triggers could be a promising area for new AI/ML research papers. According to the reviewed literature, the integration of process enhancement and process improvement has demonstrated superior performance and applicability within the industry. This combined approach ensures the integrity of the process evaluation and the implementation of corrective actions, thereby fostering a cohesive and comprehensive strategy. By simultaneously focusing on enhancing existing processes and improving their overall efficiency, organizations can achieve more consistent and reliable outcomes. Furthermore, this integrated approach presents a promising avenue for future research, particularly to explore the impacts and potential benefits of such a dual focus.
\section{Conclusions and Research Gaps}

In recent years, there has been a growing number of AI-BPM research papers addressing challenges at an enterprise level. This study introduces an application-based approach to explore papers that focus on AI-tailored solutions for identifying and responding to business process management opportunities (\emph{e.g.}, predictive business process monitoring, and automated improvement). Exploring this field allows both industry and academia to uncover innovative approaches that can enhance business initiatives through automation, in higher quality and efficiency, and fostering greater agility. This analysis focuses on two fundamental pillars within BPM: Process enhancement and Process improvement, examining their interconnection to demonstrate their practicality and relevance. This approach aims to proactively identify and resolve bottlenecks on a regular basis within a short timeframe or continuously enhance overall operations over the long term. 

Advancements in technology have enabled the processing of complex event logs from business process management systems. Analyzing research papers reveals that the influence of AI and machine learning progress extends to incorporating AI-driven decision making within the context of automation and assist in organizing workflow and automating information-intensive back-office processes. The majority of AI applications currently in use primarily aim to enhance resource management, eliminating constraints and inefficiencies inherent in the process. This typically involves making instant decisions during the process rather than depending on human involvement or less efficient rule-based systems. The goal is to prevent bottlenecks in a way that eliminates the need to wait for human analysis and decisions, or avoid predictive errors that may occur by the rule-based system while waiting for supervision. Conversely, this approach can save organizations substantial costs through predictive monitoring and evaluation (Process Enhancement), and real-time improvement (Process Improvement).

In the BPM research field, both process enhancement and process improvement have been active areas of study. However, there is potential in combining and merging these approaches to automate the well-known BPM lifecycle within organizations. Researchers have been exploring various machine learning methods such as classification, clustering, decision trees, and DNN to assess and enhance processes simultaneously, with simulation and root-cause analysis being two prominent methods utilized.

It is important to note that this is a relatively new area of research. However, compared to treating process enhancement and process improvement as separate objectives, a unified approach offers advantages in terms of scope, focus, and methodology. Based on findings in the literature, having a cohesive scope and objective can lead to better interpretability, accuracy, integrity, and a deeper understanding of business needs and stakeholder requirements. 

One major challenge in business process management is the lack of robust datasets for testing and the absence of adequate validation methods. Additionally, many information systems within businesses struggle to uncover hidden processes, which hinders advancements in automating the BPM lifecycle. However, most recent approaches have focused on supervised learning. Due to data availability issues, there is a significant need to consider unsupervised learning and deep learning methods. Additionally, analysts can benefit from few-shot or zero-shot learning to handle datasets with incomplete labels or traces, a common issue for most event logs. These models have demonstrated their effectiveness in handling such tasks in other contexts. Noteworthy, transparency and explainability are crucial for these models, as businesses value explainable models to understand the contribution of each feature. Without this, the risk of failure increases in BPM projects. A potential solution to the issue of lacking data is the use of Generative Adversarial Networks (GANs), Transformers, and Autoencoders to generate synthetic datasets.

In summary, by introducing the diverse approaches of AI and machine learning in BPM, we answered the research questions. There is a trend in developing machine learning techniques to enhance usability, accuracy, and interpretability, BPM also needs to be adapted to these advancements to avoid being underused. Challenges such as process drift analysis and remaining time prediction, etc. have been major issues in BPM and AI provides a wide range of methods to avoid occasional human errors and a lack of transparency. Generally speaking, process optimization still does not support full automation. To be more specific, automated solutions covering the enhancement and improvement of processes are in demand. Particularly, generative AI could be a valuable opportunity to automate process redesign toward the improved models.

\section{Acknowledgements}
We acknowledge the support of the Natural Sciences and Engineering Research Council of Canada (NSERC), [funding reference number ALLRP 561264 - 21]. 
Cette recherche a été financée par le Conseil de recherches en sciences naturelles et en génie du Canada (CRSNG), [numéro de référence ALLRP 561264 - 21].
\bibliographystyle{agsm.bst}

\bibliography{References/pi, References/pe, References/pepi, References/Advocate, References/References,References/proposal }
\end{document}